\documentclass[sigconf,authorversion,nonacm]{acmart}
\usepackage{algorithm}
\usepackage{algorithmic}
\usepackage{amsmath}
\usepackage{booktabs}
\usepackage{tabularx}
\usepackage{multirow}
\usepackage{algorithm}
\usepackage{algorithmic}
\usepackage{times}
\usepackage{helvet}
\usepackage{courier}
\usepackage{xcolor}
\usepackage{makecell}
\usepackage{subfig}

\AtBeginDocument{%
  }

\begin{document}

\title{A Novel FACS-Aligned Anatomical Text Description Paradigm for Fine-Grained Facial Behavior Synthesis}

\author{Jiahe Wang}
\email{wangjiahe317@mail.ustc.edu.cn}
\affiliation{%
  \institution{University of Science and Technology of China}
  \city{Hefei}
  \country{China}
}

\author{Cong Liang}
\affiliation{%
  \institution{University of Science and Technology of China}
  \city{Hefei}
  \country{China}
  }
\email{lc150303@mail.ustc.edu.cn}

\author{Xuandong Huang}
\affiliation{%
  \institution{University of Science and Technology of China}
  \city{Hefei}
  \country{China}
  }
\email{xuandong@mail.ustc.edu.cn}

\author{Yuxin Wang}
\affiliation{%
  \institution{University of Science and Technology of China}
  \city{Hefei}
  \country{China}
  }
\email{wyx2020@mail.ustc.edu.cn}

\author{Xin Yun}
\affiliation{%
  \institution{University of Science and Technology of China}
  \city{Hefei}
  \country{China}
  }
\email{yx9329@mail.ustc.edu.cn}

\author{Yi Wu}
\affiliation{%
  \institution{University of Science and Technology of China}
  \city{Hefei}
  \country{China}
  }
\email{wy221711@mail.ustc.edu.cn}

\author{Yanan Chang}
\affiliation{%
  \institution{University of Science and Technology of China}
  \city{Hefei}
  \country{China}
  }

\author{Shangfei Wang}
\authornote{Corresponding author}
\affiliation{%
  \institution{University of Science and Technology of China}
  \city{Hefei}
  \country{China}
  }
\email{sfwang@ustc.edu.cn}

\begin{abstract}
Facial behavior constitutes the primary medium of human nonverbal communication. Existing synthesis methods predominantly follow two paradigms: coarse emotion category labels or one-hot Action Unit (AU) vectors from the Facial Action Coding System (FACS). Neither paradigm reliably renders fine-grained facial behaviors nor resolves anatomically implausible artifacts caused by conflicting AUs. Therefore, we propose a novel task paradigm: anatomically grounded facial behavior synthesis from FACS-based AU descriptions. This paradigm explicitly encodes FACS-defined muscle movement rules, inter-AU interactions, and conflict resolution mechanisms into natural language control signals. To enable systematic research, we develop a dynamic AU text processor, a FACS rule-based module that converts raw AU annotations into anatomically consistent natural language descriptions. Using this processor, we construct BP4D-AUText, the first large-scale text-image paired dataset for fine-grained facial behavior synthesis, comprising over 302K high-quality samples. Given that existing general semantic consistency metrics cannot capture the alignment between anatomical facial descriptions and synthesized muscle movements, we propose the Alignment Accuracy of AU Probability Distributions (AAAD), a task-specific metric that quantifies semantic consistency. Finally, we design VQ-AUFace, a robust baseline framework incorporating anatomical priors and progressive cross-modal alignment, to validate the paradigm. Extensive quantitative experiments and user studies demonstrate the paradigm significantly outperforms state-of-the-art methods, particularly in challenging conflicting AU scenarios, achieving superior anatomical fidelity, semantic consistency, and visual quality.
\end{abstract}


\maketitle

\section{Introduction}
\label{sec:intro}

\begin{figure}[t]
    \centering
    \includegraphics[width=\linewidth]{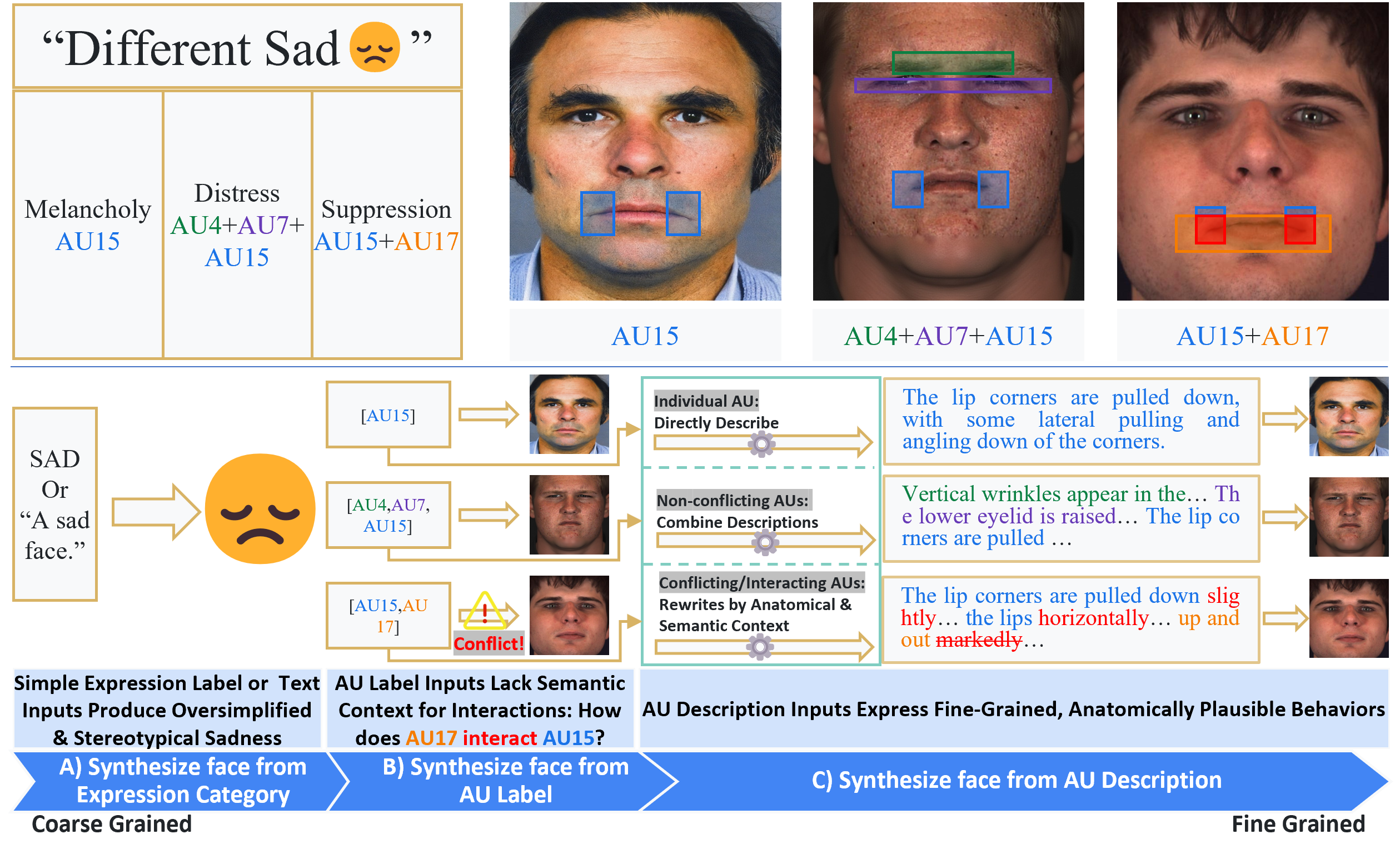}
    \caption{The upper left panel demonstrates three distinct subtypes of sadness and their corresponding AU combinations. The upper right panel visualizes facial regions affected by different AUs, with red marking conflicts caused by simultaneous activation of antagonistic AUs. The lower panel compares three paradigms: (A) Expression category-driven face synthesis; (B) AU label-driven face synthesis; (C) Our proposed AU description-driven face synthesis, with our Dynamic AU Text Processor highlighted in the green box. Full AU descriptions are provided in Appendix~\ref{sec:autext}.}
    \label{fig:interactions}
\end{figure}

Facial behavior synthesis is a core foundational task in multimedia, digital human creation, and human-computer interaction. Facial behavior conveys the full spectrum of human emotions and intentions through the coordinated activation of over 40 facial muscles, with psychological studies estimating that humans can produce approximately 3,000 distinct, meaningful facial configurations~\cite{practicalpsychology2021}. Despite significant progress in controllable face synthesis, existing methods still struggle to capture the subtle yet behaviorally critical variations in human facial expressions.

Existing facial behavior synthesis methods follow two dominant paradigms, both of which have fundamental limitations. The first paradigm relies on coarse emotion category labels, such as the eight basic expressions. We call them \textit{Paradigm 1: Coarse text/attribute label-driven text-to-face methods}. As shown in Figure~\ref{fig:interactions}A, even the single label "sad" encompasses multiple distinct psychological states: the calm detachment of melancholy, the anxious sorrow of distress, and the restrained tension of suppression~\cite{williams2024keatsian,kang2025risk,hesser2013costs}. Conventional text-to-face models in this paradigm~\cite{Sun_Li_Wang_Zhao_Sun_2021,Yan_Zhang_Wang_Cheng_Yu_Fu_2023,Wang_Bai_Wang_Qin_Chen_2024,he2024uniportraitunifiedframeworkidentitypreserving} use generic emotion prompts or attribute labels as control signals, which lack the granularity to capture these nuanced expression variations. These methods collapse the inherent diversity of facial behavior into stereotyped outputs, failing to disentangle fine-grained expression details during synthesis.

The second paradigm uses the Facial Action Coding System (FACS) for more precise control, encoding facial movements as discrete one-hot Action Unit (AU) vectors~\cite{Facial_Action_Coding_System}. AUs are anatomically grounded in individual facial muscle movements, offering higher precision than coarse emotion labels. We call them \textit{Paradigm 2: One-hot AU label-driven facial behavior synthesis methods}. However, existing methods in this paradigm~\cite{pumarola2018ganimation,https://doi.org/10.1111/cgf.14202,wu2020cascade} have a critical flaw: they model compound expressions as simple linear superpositions of individual AUs. This linear assumption fails completely when handling conflicting AUs, defined as antagonistic activations of the same facial muscle group with opposing actions. As illustrated in Figure~\ref{fig:interactions}B, AU15 and AU17 form a typical conflicting pair: both act on the mouth corners, with AU15 pulling them downward and AU17 lifting them upward. When co-activated, their physiological interaction should produce a subtle downward pull of the mouth corners with tightened lips, reflecting natural muscle competition~\cite{Facial_Action_Coding_System}. A naive linear superposition, however, often generates anatomically implausible artifacts, such as simultaneous upward and downward motion of the mouth corners, or inconsistent muscle activation patterns. These artifacts arise because label-driven models lack the mechanism to resolve inherent biomechanical incompatibilities between co-occurring AUs.

To address the limitations of both existing paradigms, we propose a novel task paradigm: anatomical description-driven fine-grained facial behavior synthesis. This paradigm uses FACS-based natural language descriptions of AUs as the core control signal for synthesis. Unlike discrete labels, AU Descriptions explicitly encode FACS-defined muscle movement rules, inter-AU interactions, and conflict resolution mechanisms into semantically grounded natural language. This breaks the linearity constraint of existing label-driven methods, enabling the synthesis of fine-grained, anatomically plausible facial behaviors that cannot be achieved by either existing paradigm.

To enable systematic research on this new task, we first address the absence of dedicated training and evaluation data. We construct BP4D-AUText, the first large-scale text-image paired dataset tailored for anatomical description-driven facial behavior synthesis. Built from the BP4D and BP4D+ datasets via our proposed Dynamic AU Text Processor, the dataset contains over 302K high-quality samples with FACS-aligned AU descriptions, with dedicated handling for conflicting AU combinations. As illustrated in Figure~\ref{fig:interactions}C, our Dynamic AU Text Processor generates AU Descriptions following three core principles: (1) individual AUs are rendered using their canonical FACS definitions; (2) non-conflicting AUs are combined via semantically coherent concatenation; and (3) conflicting or functionally interacting AUs are rewritten into integrated phrases that reflect their joint physiological effect. This design ensures the input descriptions provide both precise anatomical guidance and rich contextual semantics for the synthesis task.

To standardize the evaluation of the new task, we further propose AAAD (Alignment Accuracy of AU Probability Distributions), a novel metric that quantifies the anatomical semantic consistency between input AU descriptions and synthesized facial images. AAAD aligns the AU activation probability distributions predicted from the input text and the generated image, providing a principled, quantitative way to assess whether synthesized expressions accurately reflect the intended anatomical descriptions.

To validate the feasibility of our proposed paradigm, we design VQ-AUFace, a baseline framework for the new task. VQ-AUFace incorporates two key designs tailored to the requirements of anatomical description-driven synthesis: an anatomical prior-driven facial codec that enforces biomechanical consistency in generated faces, and a progressive cross-modal alignment module that facilitates the precise mapping of conflict-resolved semantic descriptions by aligning textual descriptions with visual facial features across spatial and semantic levels. Extensive experiments demonstrate that this baseline effectively generates anatomically plausible, behaviorally coherent facial behaviors from AU Descriptions, especially in challenging scenarios with conflicting AUs.

In summary, our key contributions are as follows:

\noindent 1. We propose a novel anatomical description-driven paradigm for facial behavior synthesis, defining a new research direction that overcomes the fundamental limitations of existing label-driven methods.

\noindent 2. We present \textbf{BP4D-AUText}, the first large-scale text-image paired dataset tailored for fine-grained facial behavior synthesis, with dedicated handling of conflicting AU combinations.

\noindent 3. We propose \textbf{AAAD}, a novel evaluation metric designed to quantify the anatomical semantic consistency between input AU descriptions and synthesized facial behaviors.

\noindent 4. We develop \textbf{VQ-AUFace}, a reference implementation tailored for this task, to validate the feasibility of our proposed paradigm.

We hope this work, with its novel task formulation, standardized dataset and evaluation protocol, and validated baseline framework, will catalyze future research on linguistically grounded, anatomically faithful facial behavior synthesis.

\section{Related works}

\subsection{Synthesize Facial Behavior from AU Labels}
FACS\cite{Facial_Action_Coding_System} describes facial actions as basic muscle movements and is used in nonverbal communication, emotion, and social studies. Albert et al. first synthesized facial behavior with AU vectors and an attention mechanism\cite{pumarola2018ganimation}. Zhao et al. improved this method by enhancing a GAN framework with an AU intensity analysis network for greater anatomical plausibility\cite{https://doi.org/10.1111/cgf.14202}. Wu et al. proposed Cascade EF-GAN\cite{wu2020cascade}, a progressive synthesis approach that minimizes artifacts and blurriness around the eyes, nose, and mouth. However, these methods rely solely on discrete AU labels, which discard the rich anatomical semantics and interaction rules defined in the FACS manual, making it impossible to model non-linear relationships and conflicts between co-occurring AUs. In contrast, we are the first to use the AU Description, derived from FACS definitions and explicitly handling conflicting AU combinations, for face synthesis. Our approach thoroughly capitalizes on the nuanced semantics in facial behaviors to synthesize complex facial behaviors that are fine-grained, anatomically plausible, and highly consistent with the input text.

\subsection{Synthesize Facial Behavior from Expression Categories}

Some studies label facial behavior into eight expression categories, i.e., happiness, sadness, surprise, anger, contempt, disgust, fear, and neutral, to control facial image synthesis using contour maps\cite{huang2017dyadgan}, multi-domain translation\cite{choi2018stargan}, or masks\cite{gu2019mask}. However, advances in natural language processing have shown that fixed labels are limited\cite{Sun_Deng_Li_Sun_Ren_Sun}. As a result, text-to-face methods now convert labels to text and combine them with additional facial attribute descriptions for more flexible synthesis. Earlier approaches employed custom text encoders such as bidirectional LSTMs\cite{Chen_Qing_He_Luo_Xu_2019}, skip-thought architectures\cite{Nasir_Jha_Grover_Yu_Kumar_Shah_2019}, semantic embedding frameworks\cite{Sun_Li_Wang_Zhao_Sun_2021}, and visual-linguistic similarity modules\cite{Xia_Yang_Xue_Wu_2021}, but were limited by simple design and small training datasets. With the advent of CLIP\cite{Radford_Kim_Hallacy_2021}, researchers enhanced text encoding—for instance, combining CLIP with StyleGAN\cite{Sun_Deng_Li_Sun_Ren_Sun} and attribute memory modules\cite{10380501}—although precise control over complex facial behaviors remains challenging. Recent T2F methods using diffusion models, VQVAE, etc., include FaceStudio\cite{Yan_Zhang_Wang_Cheng_Yu_Fu_2023} and InstantID\cite{Wang_Bai_Wang_Qin_Chen_2024}, which extract global facial features with models like ArcFace\cite{Deng_2022} and Antelopev2\cite{deepinsight2025}, but often lack spatial detail for facial deformations. FlashFace\cite{zhang2024flashfacehumanimagepersonalization} leverages ReferenceNet\cite{zhang2023referenceonlycontrolnet} but requires large, multi-image datasets. More advanced works like IPAdapter\cite{ye2023ipadaptertextcompatibleimage}, Infinite-ID\cite{Wu_Li_Zheng_Wang_Li}, and UniPortrait\cite{he2024uniportraitunifiedframeworkidentitypreserving} use CLIP’s image encoder and FaceID encoder to capture local facial features. However, none incorporate facial anatomical prior knowledge, resulting in anatomically implausible synthesized facial behaviors. Furthermore, these methods cannot accept AU Descriptions as input, as the textual length and fine-grained semantic of AU Descriptions exceed the encoding capacity limits of CLIP Text Encoders. 

To address these limitations, we introduce an anatomically inspired loss function into the T2F framework, enhancing the anatomical plausibility of synthesized facial behaviors. Additionally, we replace the commonly used CLIP Text Encoder with a pre-trained language model to accommodate the length of AU Descriptions and extract more granular semantic features. These textual features are then aligned with facial behaviors in images through a progressive alignment-fusion strategy, achieving superior semantic consistency in facial behavior synthesis.

\subsection{Text-to-Face Synthesis Dataset}

As shown in Appendix~\ref{sec:t2fdataset}, text-to-face synthesis methods aim to generate facial images consistent with input text, which necessitates the use of datasets containing high-quality text–face pairs. Several datasets have been proposed to support this task. Face2Text~\cite{gatt2018face2text}, the first dataset with textual descriptions for face generation, includes 400 images annotated with physical attributes and emotions such as “serious expression” or “appears happy or sad.” SCUText2face~\cite{Chen_Qing_He_Luo_Xu_2019} contains 1,000 images and provides manual annotations covering age, gender, hair, skin, and lip color, eye shape, and smile. The dataset introduced by Nasir et al. for Text2FaceGAN~\cite{Nasir_Jha_Grover_Yu_Kumar_Shah_2019} consists of 10,000 images from CelebA~\cite{Liu_Luo_Wang_Tang_2015}, each paired with short attribute-based descriptions focusing on face shape, facial hair, hair, facial features, appearance, and accessories, though facial behaviors are limited mainly to smiling. CelebAText-HQ~\cite{Sun_Li_Wang_Zhao_Sun_2021} includes 15,010 frontal faces from CelebAMask-HQ\cite{CelebAMask-HQ}, with manually written descriptions detailing gender, age, face shape, expression, lips, nose, ears, skin, hairstyle, hair color, facial hair, eyes, and accessories. Multi-Modal CelebA-HQ~\cite{Xia_Yang_Xue_Wu_2021} extends CelebA-HQ\cite{karras2018progressive} by providing ten distinct captions per image, covering attributes such as age, gender, and hairstyle, yet still omitting detailed facial behaviors. FFText-HQ~\cite{10380501} augments FFHQ\cite{Karras_Laine_Aila_2019} with one sentence per image describing hair, eyes, mouth, nose, face structure, makeup, accessories, expression, age, and gender.
A key limitation of existing T2F datasets is their predominant focus on static facial attributes, with very limited coverage of dynamic facial behaviors. To address this gap, we developed the Dynamic AU Text Processor based on FACS and constructed BP4D-AUText, the first large-scale dataset specifically designed for complex facial behavior synthesis. This dataset paves the way for more advanced and anatomically grounded facial expression generation.
\section{Dataset Construction}
This section details the construction process of the BP4D-AUText dataset. We first describe the selection criteria for raw data sources. Subsequently, we introduce the dynamic AU text processor used for annotation generation. We then explain the protocol for reference face selection. Finally, we present a statistical analysis of the dataset composition.
\subsection{Data Source}
Prominent AU datasets include the BP4D series and DISFA \cite{Mavadati_Mahoor_Bartlett_Trinh_Cohn_2013}. The BP4D series comprises BP4D \cite{Zhang_Yin_Cohn_Canavan_Reale_Horowitz_Liu_Girard_2014} and BP4D+ \cite{Zhang_Girard_Wu_Zhang_Liu_Ciftci_Canavan_Reale_Horowitz_Yang_et_al._2016}, and the combined number of images in these datasets far exceeds that of DISFA and other AU datasets. Therefore, we employ the BP4D series as the primary data source. We construct a large-scale dataset named \textbf{BP4D-AUText} by processing 302,169 facial images from the BP4D and BP4D+ datasets. These images are carefully selected to ensure each contains at least one activated AU. Each facial image in the dataset is accompanied by a fine-grained textual description detailing the facial behaviors and a corresponding reference face image.
\subsection{Dynamic AU Text Processor}
We build on the semantic descriptions of 15 AUs summarized by \citet{Yang_Yin_Zhou_Gu_2021} from the FACS manual, adopting them as the textual annotations for individual AUs. However, we observe that when multiple AUs co-occur, their interactions can alter the resulting facial muscle movements in ways that are not captured by simply concatenating individual AU descriptions. To accurately represent these combined effects, we consult the documented Appearance Changes for specific AU combinations in the FACS Manual by Ekman \cite{Facial_Action_Coding_System}. By integrating multiple individual AU descriptions, we construct semantically and anatomically consistent textual descriptions for 26 interacting AU combinations present in the dataset. To enhance linguistic diversity and improve model generalization, we employ the large language model GLM to generate four paraphrased variants for each original description, resulting in five semantically equivalent but lexically distinct descriptions per instance. We manually verify a subset of the machine-generated texts to ensure semantic consistency with the original annotations. In total, the dataset covers 38 AU constructs, consisting of 12 individual AUs and 26 interacting combinations, each associated with five textual descriptions. We further introduce an algorithm to determine the correct compositional structure of activated AUs. The complete set of AU descriptions and the text selection algorithm are provided in Appendix \ref{sec:fullautext} and Appendix \ref{sec:textusagealorithm}, respectively.
\subsection{Reference Face Selection}
We randomly select an image with no AUs activated as the reference face for each individual. In cases where no image exists for an individual with all AUs inactive, we select the image with the smallest index as the reference face for that individual.
\subsection{Data Statistics}
\begin{figure}[htbp]
\centering
\includegraphics[width=\linewidth]{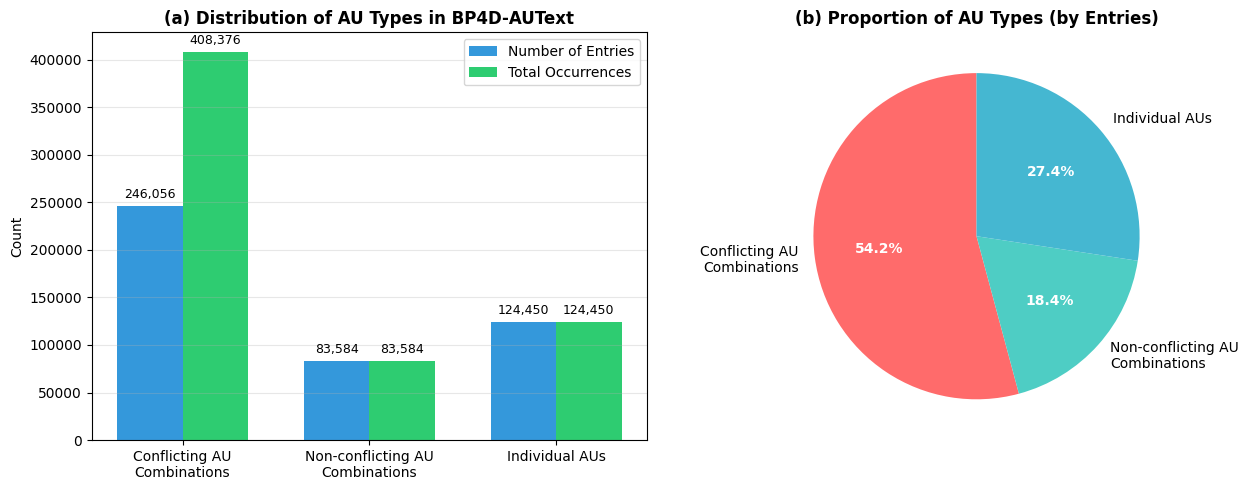}
\caption{Statistical Analysis of AU Distribution in BP4D-AUText Dataset. (a) Bar chart comparing the number of entries and total occurrences across three AU categories, highlighting the prevalence of conflicting combinations. (b) Proportional distribution of AU types based on entry count, demonstrating the dominance of complex facial behavior patterns in spontaneous expressions.}
\label{fig:dataset_stats}
\end{figure}

The statistical distribution of AU types in the BP4D-AUText dataset, illustrated in Figure~\ref{fig:dataset_stats}, underscores the critical challenge of modeling complex facial interactions. Conflicting AU combinations dominate the dataset, comprising 81.4\% of all entries (246,056 images) and 408,376 total occurrences. The fact that the number of occurrences vastly exceeds the number of entries indicates that images frequently contain multiple conflicting pairs, revealing the inherent non-linearity of spontaneous expressions. This prevalence highlights that the ability to resolve such conflicts is not a corner case but a fundamental requirement for anatomically plausible synthesis, as conventional linear modeling assumptions are inadequate for the majority of real-world facial behaviors. The composition of the dataset thus provides a realistic and challenging benchmark for advancing the field.
\section{Task Formulation and Baseline Framework}
\begin{figure}[ht]
    \centering
    \includegraphics[width=\linewidth]{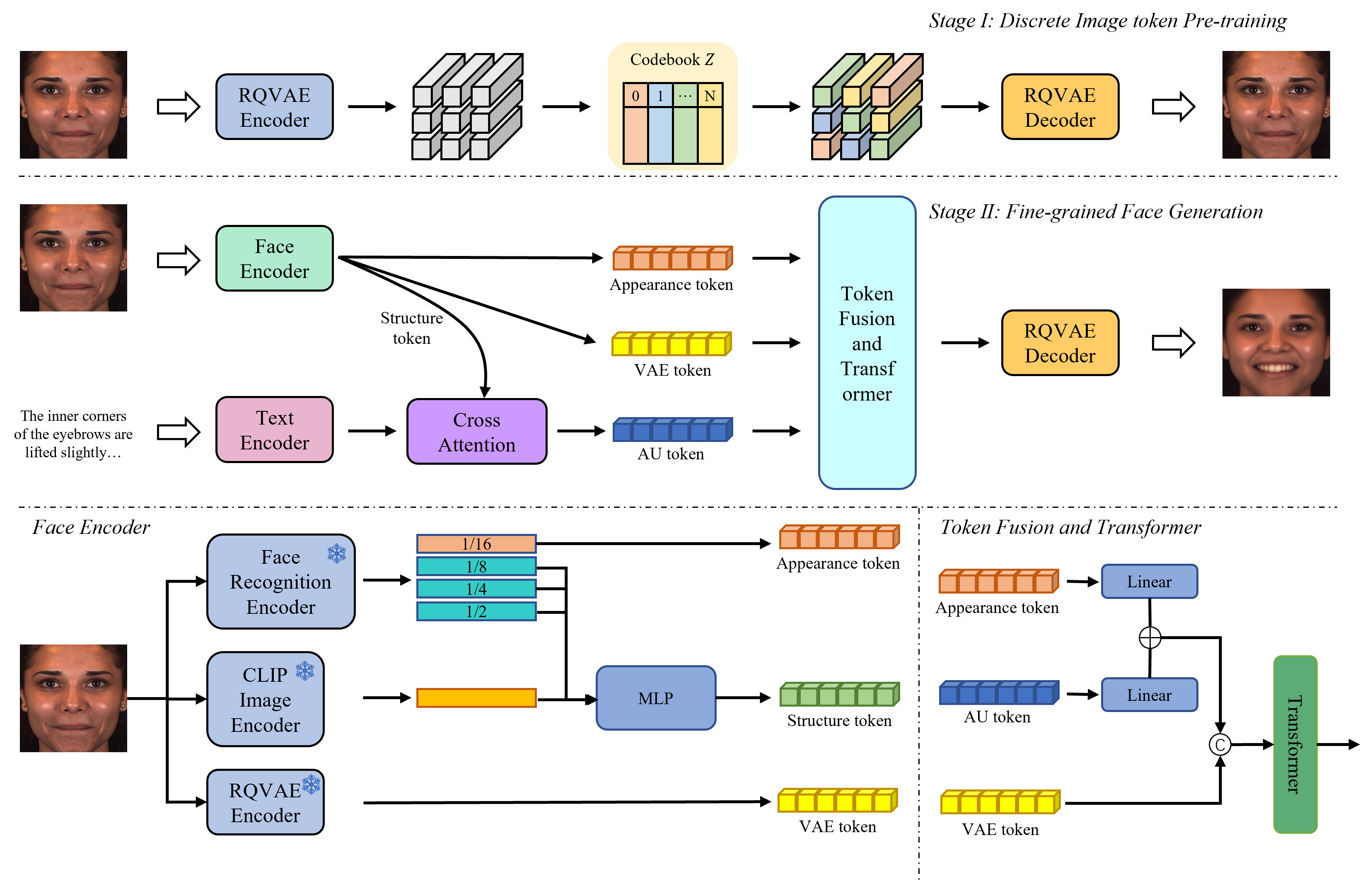}
    \caption{Overview of the VQ-AUFace baseline framework. The framework addresses the proposed facial behavior synthesis task by generating facial images conditioned on a reference face and FACS-aligned AU descriptions. The Face Encoder extracts features that preserve identity and structure, whereas the Text Encoder processes AU descriptions. Cross-attention between structural and text tokens yields AU-specific guidance, which is fused with appearance and VAE tokens to guide the Transformer-based generator.}
    \label{fig:framework}
\end{figure}

We formally define the proposed facial behavior synthesis task. The task involves two primary inputs: a reference face image $I_R$ preserving the identity of the target subject, and a FACS-aligned AU description text $T_{AU}$ encoding anatomical rules of facial muscle movements, inter-AU interactions, and mechanisms for resolving conflicting AUs. The output is a photorealistic facial image $I_G$, which must retain the identity of $I_R$ while ensuring the facial behavior aligns fully with the anatomical description in $T_{AU}$. The primary objective is to maximize anatomical semantic consistency between $I_G$ and $T_{AU}$, while maintaining visual fidelity and identity preservation in the generated image.

Prior work by Zhang et al. \cite{zhang2023generating} indicates that, compared to continuous generative models (e.g., diffusion models, GANs), VQ-VAE offers a more robust structural prior for mapping granular linguistic descriptions to anatomical actions. Consequently, we establish the VQ-AUFace framework as a baseline to validate the feasibility of the proposed task. This framework incorporates facial anatomical priors and fine-grained textual guidance, comprising an overall pipeline of two sequential stages: Discrete Image Token Pre-training and Fine-grained Face Generation. Stage I learns discrete facial representations conforming to anatomical structures, thereby providing a robust token space for Stage II. Stage II leverages these representations to perform conditional generation guided by AU descriptions. The key components of the baseline framework are described in the following sections.

\subsection{Stage I: Discrete Image Token Pre-training}
This stage focuses on learning discrete facial image representations that conform to FACS anatomical structures, to support precise generation of fine-grained facial behaviors required by the proposed task. We employ a Residual VQGAN for image discretization, whose architecture draws inspiration from VQGAN\cite{Esser_Rombach_Ommer_2021} and RQ-VAE\cite{lee2022autoregressive}. We adapt its quantization layer to handle multi-scale facial features, ensuring the model can reconstruct anatomically plausible facial structures from discrete tokens after pre-training.

To train the encoder $E$ and decoder $G$ of the Residual VQGAN, we employ gradient descent to optimize the loss $\mathcal{L}_1=\mathcal{L}_{recon}+\mathcal{L}_{commit}$. The reconstruction loss $\mathcal{L}_{recon}$ and the commitment loss $\mathcal{L}_{commit}$ are defined as follows:
\begin{gather}
    \mathcal{L}_{recon} = \left| I-\hat I \right|\\
    \mathcal{L}_{commit} = \frac{1}{N}\sum_{i =1}^{N} \mathcal{L}^{(i)}_{quantizer}
\end{gather}
where $N$ denotes the number of quantizers, and the commitment loss for each quantizer $\mathcal{L}^{(i)}_{quantizer}$ is computed using the same method as prior RQ-VAE research\cite{lee2022autoregressive}.

For the adversarial part of the Residual VQGAN, we alternately optimize the losses $\mathcal{L}_{gen}$ and $\mathcal{L}_{discr}$ using gradient descent. The generator loss $\mathcal{L}_{gen}$ and the discriminator loss $\mathcal{L}_{discr}$ are defined as follows:
\begin{gather}
    \mathcal{L}_{{gen}} = -\mathbb{E}_{\text{fake}} \left[ D(\text{fake}) \right]
    \\[1mm]
    \mathcal{L}_{{discr}} = \mathbb{E}_{\text{real}} \left[ \max(0, 1 - D(\text{real})) \right]+\\\mathbb{E}_{\text{fake}} \left[ \max(0, 1 + D(\text{fake})) \right]\notag
\end{gather}
where $D(\text{real})$ denotes the output of the discriminator for real samples, $D(\text{fake})$ for generated samples, and $\mathbb{E}_{\text{real}}$ and $\mathbb{E}_{\text{fake}}$ represent the expectations over real and generated samples respectively.

We employ facial anatomical loss and perceptual loss to drive the model to generate faces that conform to FACS anatomical constraints, which fully matches the core goal of the proposed task. Specifically, the facial anatomical loss and perceptual loss are defined as follows:
\begin{gather}
\small 
    \mathcal{L}_{anatomical} = \left| \text{MEF}(I)-\text{MEF}(\hat I) \right|^2
    \\
    \mathcal{L}_{perceptual} = \left| \text{vgg}(I)-\text{vgg}(\hat I) \right|^2
\end{gather}
where $\text{vgg}(\cdot)$ denotes the feature output of the VGG16 model\cite{simonyan2015deep} after removing its last two layers, and $\text{MEF}(\cdot)$ represents the feature output of the MEFARG method in ME-GraphAU\cite{luo2022learning,song2022gratis} after discarding its final layer.

Considering that the reconstructed images from the decoder contain no meaningful facial structures in the early training stage, we do not compute the facial anatomical loss at the beginning of the training phase to conserve computational resources and enhance training stability. We introduce the facial anatomical loss into the optimization process once the reconstructed images exhibit basic facial structures.

\subsection{Stage II: Fine-grained Face Generation}
This stage implements the core inference logic of the proposed task, leveraging AU Descriptions as input to guide fine-grained facial behavior synthesis. The stage comprises three key components: the Face Encoder, the Text Encoder with AU Cross Attention, and the Token Fusion with Transformer.

First, the Face Encoder extracts three complementary feature types from the reference face $I_R$. Appearance tokens $T_A$ capture identity-related attributes such as face shape and skin texture. Structure tokens $T_S$ encode spatial information related to facial muscles and texture layout. VAE tokens $T_V$ provide low-level image features that are essential for high-fidelity reconstruction. These features are derived through a combination of pre-trained encoders, including a Face Recognition Encoder, a CLIP Image Encoder, and an RQVAE Encoder. This ensemble of encoders collectively ensures robust identity preservation and accurate spatial alignment throughout the synthesis process.

Second, the Text Encoder and AU Cross Attention module processes textual guidance. AU Descriptions are encoded into text tokens $T_T$ using T5-large. We select T5-large to replace the widely used CLIP Text Encoder, to adapt to the long text length and fine-grained anatomical semantics of AU Descriptions required by the proposed task, and to resolve the encoding capacity bottleneck of existing text-to-face methods. To align textual guidance with facial structure, we compute cross-attention between $T_T$ and $T_S$, generating AU tokens $T_{AU}$ that encode spatially coherent muscle activation patterns. This step ensures the model can accurately align with the complex AU interactions explicitly encoded in the natural language descriptions.

Third, the Token Fusion and Transformer module generates the final output. The final context for generation is formed by fusing $T_A$, $T_{AU}$, and $T_V$. A weighted sum of $T_A$ and $T_{AU}$ is concatenated with $T_V$, and the resulting context is fed into a MaskGitTransformer to generate discrete image tokens $T_I$. The Residual VQGAN decoder then reconstructs the final image $I_G$.

\section{Experiments}
This section systematically validates the effectiveness of our proposed anatomical description-driven facial behavior synthesis paradigm. All experiments are conducted on the BP4D-AUText dataset constructed for the proposed new task. Detailed model implementation configurations are provided in Appendix \ref{sec:impofvqauface}.

\subsection{Evaluation Metrics}
We employ two categories of metrics to comprehensively evaluate the performance of all methods, covering both image quality and task-specific anatomical semantic consistency. For image quality assessment, we utilize four widely established metrics: Fréchet Inception Distance (FID), Kernel Inception Distance (KID), Inception Score (IS), and Learned Perceptual Image Patch Similarity (LPIPS), which quantify the visual fidelity and realism of synthesized facial images. Regarding anatomical semantic consistency, to the best of our knowledge, there is no existing semantic consistency metric specifically designed for the anatomical description-driven facial behavior synthesis task. Existing general text-image consistency metrics cannot accurately capture the alignment between FACS anatomical descriptions and facial muscle movements in synthesized images. To address this gap, we propose a novel metric termed \textbf{AAAD}, which quantifies the consistency between input text descriptions and synthesized images by aligning the probability distributions of facial AU activations across both modalities. Specifically, given the AU activation probabilities predicted from a synthesized image $P_i$ and from the input AU description text $P_t$, we compute their cosine similarity as follows:
\begin{gather}
\text{AAAD} = 
\mathrm{norm}\!\left(
    \frac{P_i^\top P_t}{\|P_i\|_2 \|P_t\|_2}
\right)
\end{gather}
where $norm(\cdot)$ denotes a normalization function that maps the cosine similarity into a standardized range using the maximum and minimum similarity values computed over the test set. Further implementation details of AAAD are provided in Appendix \ref{sec:detailsofAAAD}.

\subsection{Comparison Methods and Experimental Setup}
\begin{table}[htbp]
\centering
\setlength{\tabcolsep}{3pt}
\caption{Implementation setup of compared methods. Abbreviations: Eval. (Evaluation), Qual. (Qualitative), Ref. (Reference).}
\label{tab:method_setup}
\begin{tabular}{lccc}
\toprule
\textbf{Method} & \textbf{Training} & \textbf{Input} & \textbf{Eval.} \\
\midrule
GLM4 & None & Text & Qual. \\
MidJourney & None & Text+Ref. & Qual \\
UniPortrait & None & Text+Ref. & Both \\
AnyFace & None & Text & Both \\
SD 1.5 & Fine-tuned & Text & Both \\

GANimation & Retrained & AU+Ref. & Both \\
\bottomrule
\end{tabular}
\end{table}

We compare the proposed paradigm with six representative methods from the two existing dominant label-driven paradigms to validate the effectiveness of the anatomical description-driven framework. The first category comprises coarse text or attribute label-driven text-to-face methods, including GLM4\cite{glm2024chatglm}, MidJourney\cite{midjourney_image_generation}, AnyFace\cite{Sun_Deng_Li_Sun_Ren_Sun}, UniPortrait\cite{he2024uniportraitunifiedframeworkidentitypreserving}, and Stable Diffusion 1.5\cite{Rombach_2022_CVPR}. The second category consists of one-hot AU label-driven facial behavior synthesis methods, exemplified by the classic AU-driven method GANimation\cite{pumarola2018ganimation}. Detailed implementation descriptions are provided in Appendix~\ref{app:comparison_setup}.

All methods are evaluated on the same test set of the BP4D-AUText dataset. GLM4 and MidJourney are only included for qualitative analysis, while the other four methods are evaluated in both quantitative and qualitative analyses.

\begin{table}[htbp]
\small
\centering
\caption{Ablation results of AU textual representation versus AU label input.}
\label{tab:ablation_text_vs_label}
\begin{tabular}{cccccc}
\toprule
\textbf{Input Setting} & FID$\downarrow$ & KID$\downarrow$ & IS$\uparrow$ & LPIPS$\downarrow$ & AAAD$\uparrow$ \\
\midrule
\makecell{AU Textual\\Representation (Ours)}& 63.604 & 0.074  & 1.032 & 0.390  & \textbf{0.606} \\
\makecell{AU Label Input\\(MLP Projected)} & 77.546 & 0.091 & 1.026 & 0.470 & 0.459 \\
\bottomrule
\end{tabular}
\end{table}

\subsection{AU Text Ablation Validation}
The fundamental distinction between the proposed paradigm and existing label-driven paradigms resides in the input representation: the proposed method utilizes FACS-based anatomical AU descriptions, whereas existing methods employ discrete one-hot AU labels or coarse emotion labels. To verify the inherent superiority of the AU textual representation, a rigorous ablation study is conducted.

The complete architecture, training settings, and dataset of the VQ-AUFace baseline model are retained unchanged, and only the AU textual description input is replaced with one-hot AU labels. The one-hot labels are projected via a multi-layer perceptron to match the token length and feature dimension of the original text embedding. This controlled setting ensures that any performance difference is attributed exclusively to the input representation, rather than the model architecture or training settings.

The quantitative results are summarized in Table \ref{tab:ablation_text_vs_label}. Replacing the AU textual representation with discrete AU labels leads to drastic performance degradation across all metrics. Specifically, the semantic consistency metric AAAD drops from 0.606 to 0.459, and all image quality metrics also exhibit significant deterioration.

This result empirically validates the inherent superiority of the anatomical description-driven paradigm over the conventional AU label-driven paradigm. Unlike orthogonal one-hot label vectors that lack inherent biomechanical and anatomical priors, the AU text descriptions explicitly encode the physiological constraints of facial muscle movements into the semantic space, providing a strong structural prior for the model. This prior guides the model to synthesize anatomically plausible facial behaviors, rather than forcing the model to implicitly learn complex facial physics from limited training data.

\subsection{Quantitative and Qualitative Comparison}
Comprehensive quantitative and qualitative comparisons are conducted between the proposed paradigm and the two existing dominant label-driven paradigms to validate the overall effectiveness of the method. All experiments are conducted on the same test set of the BP4D-AUText dataset, with 2,000 randomly sampled test cases. The quantitative results are summarized in Table \ref{tab:Quantitative Results}.

\begin{table}[htbp]
\centering
\caption{Quantitative Results: Arrows indicate the direction of superiority for the corresponding metrics, with bold font denoting the optimal value and underlined font indicating the suboptimal value.}
\scalebox{0.75}{
\begin{tabularx}{1.3\linewidth}{l c c c c c} 
\toprule
\multirow{2}{*}{Methods} &\multicolumn{4}{c}{Image Quality} & Semantic Consistency \\
& FID $\downarrow$ &KID$\downarrow$ & IS$\uparrow$ & LPIPS $\downarrow$ & AAAD $\uparrow$ \\
\midrule
\multicolumn{6}{l}{\textbf{Paradigm 1: Coarse text/attribute label-driven methods}} \\
AnyFace \cite{Sun_Deng_Li_Sun_Ren_Sun} & 171.677 & 0.175 & 1.013 & 0.570 & 0.555  \\
UniPortrait \cite{he2024uniportraitunifiedframeworkidentitypreserving} & 108.159 & 0.092 & {1.043} & 0.568 & 0.588  \\
Stable Diffusion \cite{Rombach_2022_CVPR} & {126.217} & {0.157} & 1.008  & 0.475  & $\underline{0.602}$ \\
\midrule
\multicolumn{6}{l}{\textbf{Paradigm 2: One-hot AU label-driven methods}} \\
GANimation \cite{pumarola2018ganimation} & {29.388} & {0.030} & 1.031 & 0.502 & 0.591  \\
\midrule
\multicolumn{6}{l}{\textbf{Our Proposed Paradigm: Anatomical description-driven VQ-AUFace}} \\
VQ-AUFace  & 63.604 & 0.074  & 1.032 & 0.390  & \textbf{0.606}  \\
\bottomrule
\end{tabularx}
}
\label{tab:Quantitative Results}
\end{table}

\subsubsection{Quantitative Results Analysis}
The proposed anatomical description-driven paradigm achieves the best semantic consistency (AAAD) while maintaining high image quality, outperforming methods from both existing dominant paradigms.

Regarding the coarse text/attribute label-driven paradigm, AnyFace, which neglects facial spatial structural information and focuses only on static facial attributes, achieves the worst performance across all metrics. UniPortrait, which incorporates facial spatial structure modeling, achieves a relatively good AAAD score even without fine-tuning, far surpassing AnyFace. This indicates that facial structural prior knowledge is critical for facial behavior control, which is a core design of the proposed paradigm. Stable Diffusion, which is fine-tuned on the dataset, achieves the second-best AAAD score. However, it still lags behind the full model, as it lacks explicit anatomical prior modeling and cannot fully capture the fine-grained semantic information in AU descriptions.

Regarding the one-hot AU label-driven paradigm, GANimation, which uses discrete AU labels as input, fails to effectively transmit fine-grained semantic information of facial behaviors. Even with the same training data and settings as the model, its AAAD score remains lower than both Stable Diffusion and the full model. This directly demonstrates that AU textual descriptions contain far more anatomical semantic information than discrete AU labels, which is consistent with the core hypothesis and the previous paradigm validation experiment.

In addition, the method achieves the best LPIPS score, indicating that the synthesized images are most similar to real faces in terms of human visual perception. This demonstrates that the anatomical prior design in the paradigm effectively enhances the anatomical plausibility of synthesized faces, as human visual perception is highly sensitive to facial structural consistency.

\begin{figure}[htbp]
    \centering
    \includegraphics[width=\linewidth]{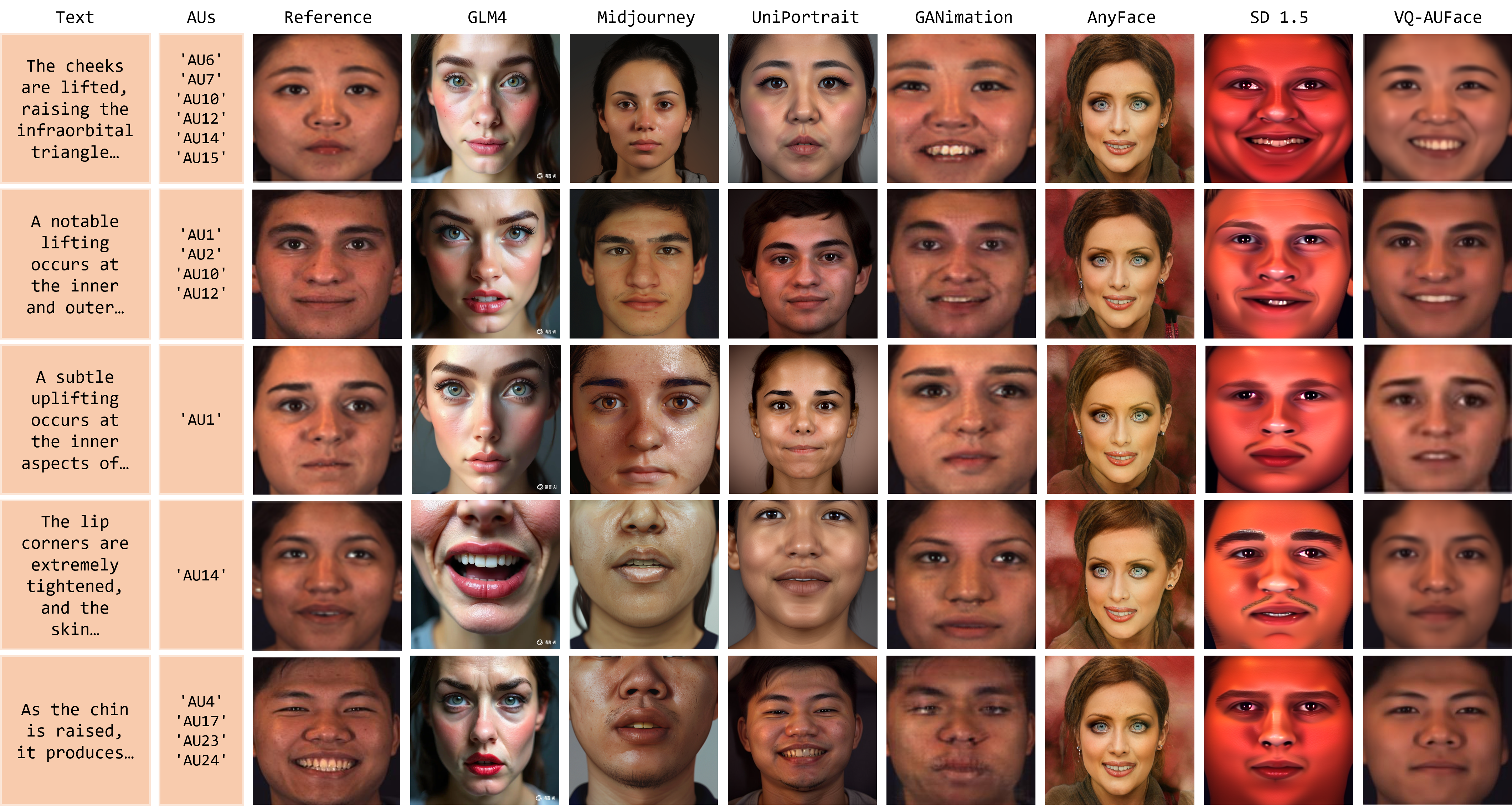}
    \caption{Comparisons with state-of-the-art methods. From left to right: Input Text, Corresponding AUs, Reference Face, and results generated by GLM4, MidJourney, UniPortrait, GANimation, AnyFace, Stable Diffusion 1.5, and VQ-AUFace (ours). Full input text sentences are provided in Appendix \ref{sec:inputtext}.}
    \label{fig:comparison}
\end{figure}

\subsubsection{Qualitative Results Analysis}
Qualitative comparisons are conducted on five representative facial behavior cases, covering single AUs, non-conflicting AU combinations, and complex conflicting AU combinations. The results are shown in Figure \ref{fig:comparison}.

Methods from the coarse label-driven paradigm (GLM4, MidJourney, UniPortrait, AnyFace) fail to generate facial images that conform to the anatomical descriptions in the input text. These methods often generate muscular details that violate facial anatomy, and even produce incomplete facial regions. For example, GLM4 only synthesizes the lower half of the face in the fourth row, and MidJourney exhibits similar truncation in the fifth row. Stable Diffusion 1.5, despite fine-tuning, only learns to control lip-related facial behaviors, and fails to accurately model eyebrow movements in most cases. It also produces consistent color bias and incorrect nasal structure rendering across all samples.

The method from the AU label-driven paradigm (GANimation) produces severe artifacts and distortions around the eyes and lips, as shown in the fifth row. Most importantly, it fails to handle conflicting AU combinations. For the conflicting AU set AU6+AU12+AU15 in the first row, GANimation cannot model the offset between the upward cheek movement and downward mouth corner pull, resulting in an anatomically implausible result with conflicting facial structures. It also fails to capture the inner eyebrow raise in the third row, due to the limited expressive ability of discrete AU labels.

In contrast, the proposed anatomical description-driven paradigm generates images with the best consistency with the input text, successfully synthesizing subtle facial behaviors and complex conflicting AU combinations as described. All synthesized faces conform to FACS anatomical constraints, with no artifacts or structural distortions, while fully preserving the identity of the reference face. More detailed qualitative results are provided in Appendix \ref{sec:morequalres}.
\begin{table}[htbp]
\centering
\setlength{\tabcolsep}{3pt}
\caption{User study results on conflicting AUs. We report the AU Conflict Handling Score (AUCHS) across 15 participants and all test combinations. A higher AUCHS indicates more anatomically faithful synthesis.}
\label{tab:conflict_ranking}
\scalebox{0.8}{
\begin{tabular}{lccccc}
\toprule
\makecell[l]{Conflicting AU\\ Combination} & VQ-AUFace & AnyFace & GANimation & UniPortrait & \makecell{Stable\\ Diffusion} \\
\midrule
AU12+15 & \textbf{4.000}  & 1.400  & 3.414  & 3.226  & 2.960   \\ 
AU1+4 & 3.506  & 1.520  & \textbf{3.814}  & 2.468  & 3.680   \\ 
AU6+12+15 & \textbf{4.242}  & 1.560  & 3.038  & 3.466  & 2.546   \\ 
AU12+15+17 & \textbf{4.360}  & 1.492  & 3.120  & 3.308  & 2.720   \\ 
AU6+12+15+17 & \textbf{4.384}  & 1.574  & 3.228  & 3.506  & 2.294   \\ 
AU1+2+4 & \textbf{4.120}  & 1.906  & 2.932  & 3.068  & 2.960   \\ 
\midrule
Mean AUCHS & \textbf{4.102}  & 1.575  & 3.258  & 3.174  & 2.860   \\
\bottomrule
\end{tabular}
}
\end{table}
\subsection{AU Conflict Handling Evaluation}
The core limitation of existing label-driven paradigms is the inability to handle conflicting AU combinations, which are the dominant pattern in real-world spontaneous facial expressions (accounting for 81.4\% of samples in the BP4D-AUText dataset). To rigorously evaluate the ability of different paradigms to handle conflicting AUs, a user study is designed under the supervision of a certified FACS expert to assess the anatomical plausibility and semantic fidelity of synthesized faces under facial muscular antagonism.

Six representative conflicting AU combinations that are frequently observed in the BP4D-AUText dataset are selected, which cover diverse antagonistic muscle movement patterns. For each combination, five models from the three paradigms (VQ-AUFace, AnyFace, GANimation, UniPortrait, Stable Diffusion) are prompted to generate five images per model, yielding 30 total images per combination. All prompts explicitly specify the target conflicting AU descriptions, with no additional conflicting muscle activations introduced, to ensure a fair comparison across all methods.

15 participants aged 18–30 are recruited to complete the ranking task. For each trial, participants rank the five generated images by their conformity to the facial action description, with expert-formulated FACS-aligned visual criteria provided to support accurate judgments without professional FACS expertise. The average ranks across trials are converted into the \textbf{AU Conflict Handling Score (AUCHS)}, with 5 points for 1st place and 1 point for 5th place. A higher AUCHS indicates superior anatomical plausibility and semantic consistency.

The user study results are summarized in Table \ref{tab:conflict_ranking}. The proposed VQ-AUFace achieves the highest mean AUCHS of 4.102, and outperforms all existing methods from the two label-driven paradigms on 5 out of 6 tested conflicting AU combinations. While GANimation from the AU label-driven paradigm performs well on the single AU1+AU4 combination, it fails to maintain stable performance on more complex multi-AU conflicting cases.

This superior performance confirms that the AU textual descriptions effectively convey the intricate dynamics of conflicting muscle actions, which cannot be captured by discrete labels. The paradigm explicitly encodes the resolution mechanism of conflicting AUs into the input text, enabling the model to synthesize physiologically coherent and semantically consistent facial configurations, which is a core breakthrough over existing label-driven paradigms.

\begin{table}[htbp]
\centering
\caption{Generalization performance on unseen AU combinations.}
\label{tab:generalization_unseen_au}
\begin{tabular}{cccccc}
\toprule
\textbf{Test Setting} & FID$\downarrow$ & KID$\downarrow$ & IS$\uparrow$ & LPIPS$\downarrow$ & AAAD$\uparrow$ \\
\midrule
\makecell{Original\\Test Set} & 63.604 & 0.074  & 1.032 & 0.390  & 0.606 \\
\makecell{Unseen AU\\Combinations }& 62.180 & 0.065 & 1.032 & 0.390 & \textbf{0.607} \\
\bottomrule
\end{tabular}
\end{table}

\subsection{Generalization Capability Evaluation}
To evaluate the generalization ability of the paradigm to unseen facial behavior descriptions, a dedicated experiment is conducted on out-of-distribution AU combinations.

An unseen test set containing 2,000 AU combinations that never appear in the BP4D-AUText dataset is constructed. This test set covers diverse facial behavior patterns, including single AU descriptions, simple multi-AU combinations, and complex conflicting AU combinations. The VQ-AUFace model is evaluated on this unseen test set, using the same evaluation metrics as the main experiments.

The evaluation results are presented in Table \ref{tab:generalization_unseen_au}. The proposed paradigm achieves even slightly better performance on the unseen AU combination test set compared to the original test set, with the AAAD score reaching 0.607, and the FID and KID metrics decreasing to 62.180 and 0.065 respectively. This result demonstrates that the paradigm has strong in-domain generalization capability, and can robustly handle novel AU textual descriptions without performance degradation. The superior generalization performance directly benefits from the semantic richness of the AU textual representation: the model learns the inherent anatomical rules of each facial action unit from the text descriptions, rather than memorizing the mapping between specific label combinations and facial movements. This enables the model to flexibly generalize to unseen combinations of facial muscle movements, which is a critical advantage over existing label-driven paradigms for practical applications.

\subsection{Component Ablation Study of the Baseline}
To validate the effectiveness of each key component in the VQ-AUFace baseline framework, a systematic ablation study is conducted. A total of 10 models are trained and divided into two groups: one group with the proposed anatomical loss, and one group without the anatomical loss. For each group, ablation studies are conducted on the three key token types from the Face Encoder: appearance tokens $T_A$, structure tokens $T_S$, and VAE tokens $T_V$. The ablation settings are as follows: without $T_S$, the AU tokens $T_{AU}$ are directly the output of the text encoder, with no cross-attention with structure tokens; without $T_A$, the AU tokens $T_{AU}$ are directly concatenated with VAE tokens $T_V$, with no weighted summation with appearance tokens; without $T_V$, the fused tokens are fed directly into the Transformer, with no concatenation with VAE tokens; and without all three tokens, the baseline model has all three token types removed.

The quantitative results are summarized in Table \ref{tab:ablation_component}. The experimental results lead to three key conclusions. First, models employing the proposed anatomical loss consistently achieve better semantic consistency (AAAD) than those without it. This indicates that the anatomical loss effectively enhances the ability of the model to model facial muscle structures, and synthesize faces that conform to FACS anatomical constraints, which is a core design of the baseline framework. Second, removing appearance tokens $T_A$ has only a minor impact on both image quality and semantic consistency, which aligns with the hypothesis that $T_A$ mainly captures high-level identity features, with limited impact on facial behavior synthesis. Third, removing structure tokens $T_S$ and VAE tokens $T_V$ leads to a noticeable decline in semantic consistency metrics. This indicates that $T_S$ and $T_V$ provide complementary facial spatial information, which is critical for aligning textual anatomical descriptions with facial muscle movements in the generated images.

Overall, the ablation study confirms that all the proposed components in the baseline framework fulfill their intended purposes, and effectively enhance the anatomical plausibility and semantic consistency of the synthesized faces.

\begin{table}[htbp]
\centering
\caption{Component ablation results of the VQ-AUFace baseline model. Arrows indicate the direction of superiority for the corresponding metrics, with bold font denoting the optimal value.}
\scalebox{0.83}{
\begin{tabularx}{1.21\linewidth}{l c c c c c} 
\toprule
\multirow{2}{*}{Model Setting} &\multicolumn{4}{c}{Image Quality} & Semantic Consistency \\
& FID $\downarrow$ &KID$\downarrow$ & IS$\uparrow$ & LPIPS $\downarrow$ & AAAD $\uparrow$ \\
\midrule
\multicolumn{6}{l}{\textbf{VQ-AUFace with Anatomical Loss}} \\
Full Model          & 63.604 & 0.074  & 1.032 & 0.390  & \textbf{0.606}  \\
w/o $T_S$ & 66.031  & 0.078 & 1.033 & 0.396 & 0.574  \\
w/o $T_A$ & 63.900 & 0.073 & 1.032 & 0.395 & 0.592  \\
w/o $T_V$ & 78.859 & 0.097 & 1.023 & 0.403 & 0.578  \\
w/o All Tokens   & 67.489 & 0.075 & 1.030 & 0.407 & 0.575  \\
\midrule
\multicolumn{6}{l}{\textbf{VQ-AUFace w/o Anatomical Loss}} \\
Full Model          & 75.412 & 0.088 & 1.040  & 0.414 & 0.599  \\
w/o $T_S$ & 60.580 & 0.069 & 1.035 & 0.408 & 0.577  \\
w/o $T_A$ & 60.056 & 0.068 & 1.032 & 0.400 & 0.578  \\
w/o $T_V$ & 86.480 & 0.105 & 1.019 & 0.384 & 0.574  \\
w/o All Tokens   & 85.379 & 0.107 & 1.011 & 0.351 & 0.571  \\
\bottomrule
\end{tabularx}
}
\label{tab:ablation_component}
\end{table}

\section{Conclusion}
In this work, we propose a novel anatomical description-driven paradigm for facial behavior synthesis, which defines a new research direction breaking the fundamental limitations of existing label-driven methods. By replacing discrete emotion labels or one-hot AU vectors with FACS-based anatomical descriptions, our paradigm explicitly encodes muscle movement rules and conflicting AU resolution mechanisms, enabling fine-grained and anatomically plausible facial behavior synthesis. To support systematic research on this new direction, we contribute three core elements: BP4D-AUText, the first large-scale dataset with FACS-aligned AU combination descriptions; AAAD, a novel metric for evaluating anatomical semantic consistency; and VQ-AUFace, a strong baseline validating the paradigm’s feasibility. 
Most importantly, our paradigm is inherently architecture-agnostic and fully compatible with all mainstream generative frameworks, including diffusion models, Transformers, and GANs. The VQ-AUFace framework serves as a validated proof-of-concept rather than a performance ceiling. We anticipate that integrating this paradigm with state-of-the-art text-to-image models such as Diffusion Transformers will yield even higher visual fidelity and more precise control, opening promising avenues for future facial behavior synthesis research. We will release the dataset to facilitate such extensions by the community.


\bibliographystyle{ACM-Reference-Format}
\bibliography{samples/sample-base}

\clearpage
\appendix
\onecolumn

\section{The Full AU Text in Figure 1}
\label{sec:autext}
This section presents the complete text in Figure \ref{fig:interactions}.

\begin{itemize}
    \item AU15: Slight downward pulling is applied to the lip corners, accompanied by some lateral pulling and a slight downturn of the corners.
    \item AU4,AU7,AU15: Vertical wrinkles appear in the glabella and the eyebrows are pulled together. The inner parts of the eye-brows are pulled down a trace on the right and slightly on the left with traces of wrinkling at the corners. The lower eyelid is raised markedly and straightened slightly, causing slight bulging, and the narrowing of the eye aperture is marked to pronounced.
    \item AU15,AU17: The lip corners are pulled down slightly with a lateral pulling and angling down, stretching the lips horizontally and changing their shape at the corners. Below the lips, the skin show pouching, bagging, or wrinkling, and the chin boss is flattened or show bulges. The chin boss is pushed up severely, causing extreme wrinkling, and the lower lip is pushed up and out, which result in a depression medially under the lower lip and an inverted-U shape of the mouth.

\end{itemize}

\section{Text-to-Face Synthesis Dataset Comparsion Table}
\label{sec:t2fdataset}
\begin{table}[htbp]
    \centering
    \scalebox{1}
    {
    \begin{tabularx}{1\linewidth}{>{\centering\arraybackslash}m{2.5cm}<{\centering}m{1.1cm}<{\centering}m{1.1cm}<{\centering}m{1.1cm}<{\centering}m{1.9cm}<{\centering}m{7.8cm}}

    \hline
        \textbf{Dataset} & \textbf{Image Count} & \textbf{Text Count} & \textbf{Text Length}& \textbf{Annotation method}&\textbf{Text Focus} \\ \hline
        Face2text & 400 & 1,400 & 1$\sim$50 &Manual annotation&physical attributes, emotion and possibly inferred attributes\\ 
        SCUText2face & 1,000 & 5,000 &10$\sim$20 &Manual annotation&  age, gender, hair color, skin color, lip color, eyes shape and smile \\ 
        Text2FaceGAN & 10,000 & 60,000 &5$\sim$15 &Generation by attribute& face shape, facial hair, hair, other facial features, appearance, and accessories \\ 
        CelebAText-HQ & 15,010 & 150,100 & 6$\sim$44&Manual annotation&gender, age, face shape, expression, lips, nose, ears, skin, hairstyle, hair color, facial hair, eyes and accessories \\ 
        Multi-Modal CelebA-HQ & 30,000 & 300,000 &5$\sim$50 &Generation by attribute& face shape, facial hair, hair, other facial features, appearance, and accessories \\ 
        FFText-HQ & 20,000 & 20,000 &100$\sim$150 &Generation by attribute&hair, eye, mouth, nose, face, makeup, accessory, expression, age, and gender\\ \hline
        {BP4D-AUText} & {302,169} & {302,169} &{50$\sim$600} & {Generation by AU}&{movements in inner corners of the eyebrows, outer part of the eyebrow, glabella, infraorbital triangle, lower eyelid, upper lip, nasolabial furrow, corners of the lips and chin }\\ \hline
    \end{tabularx}
    }
    \caption{Comparison of T2F Datasets}
    \label{tab:dataset}
\end{table}

\section{Analysis of Sadness Subcategories: Melancholy, Distress, and Suppression}

As shown in Figure \ref{fig:interactions}, taking ‘sad’ as an example, it is not a single behavioral state but a behavioral spectrum ranging from introspective melancholy to overt distress to deliberate suppression. This diversity is reflected in facial behaviors as: Melancholy\cite{williams2024keatsian}: A calm sadness accompanied by contemplation or aesthetic experience, manifested as a ‘silent sorrow’ pattern dominated by AU15. Distress\cite{kang2025risk}: A mixed state of anxiety and sadness, characterized by a ‘conflicting expression’ with the AU4+15+7 combination. Suppression\cite{hesser2013costs}: The active inhibition of emotional expression, forming a ‘restrained expression’ through AU17+15.

Clinical evidence further supports this classification. For instance, in studies on adolescent Social Anxiety Disorder, patients were clustered into two subtypes: one characterized by moderate emotional expression (akin to melancholy), and the other exhibiting high control needs and low autonomy (corresponding to suppression). These subtypes showed significant differences in functional brain connectivity patterns\cite{kang2025risk}. Similarly, children with Separation Anxiety Disorder alternately exhibit overt distress (e.g., crying, somatic discomfort) and suppressed silence (suppression) under stress, reflecting behavioral markers of distinct coping strategies\cite{ehrenreich2008separation}.

\section{Text Usage Algorithm}
\label{sec:textusagealorithm}
Different AU combinations may contain identical AUs, so allocating activated AUs to the correct combinations requires careful consideration. We posit that more complex combinations hold higher significance, and based on this premise, we designed our AU combining algorithm. The algorithm is as follows:

\begin{algorithm}[H]
\caption{Combine Activated AUs}
\label{alg:au_category_text}
\begin{algorithmic}[1]
\REQUIRE AU Label row $row$, AU list $AUList$, Combined AU list $Combined\_AUList$, dictionary $dic$
\ENSURE Generated text $txt$

\STATE Initialize $active\_aus \gets \{au \in AUList \mid row[au] = 1\}$
\STATE Initialize $res\_list \gets []$
\STATE Initialize $remove\_set \gets \emptyset$

\FORALL{$combination \in Combined\_AUList$}
    \STATE $combination\_simple \gets$ Simplify $combination$ by replacing "and" with "," and removing spaces
    \STATE $combination\_simple \gets$ Split $combination\_simple$ into a list of AUs
    \IF{All AUs in $combination\_simple$ are in $active\_aus$}
        \STATE Append $combination$ to $res\_list$
        \IF{Length of $combination\_simple > 2$}
            \FORALL{$active\_au \in combination\_simple$}
                \STATE Remove $active\_au$ from $active\_aus$
            \ENDFOR
        \ELSE
            \STATE Add all AUs in $combination\_simple$ to $remove\_set$
        \ENDIF
    \ENDIF
\ENDFOR

\STATE $active\_aus \gets \{au \in active\_aus \mid au \notin remove\_set\}$
\IF{$active\_aus$ is not empty}
    \STATE Append all $active\_aus$ to $res\_list$
\ENDIF

\STATE $txt \gets$ Concatenate random choices from $dic[au\_combination]$ for each $au\_combination \in res\_list$
\RETURN $txt$

\end{algorithmic}
\end{algorithm}

In this algorithm, ‘AU Label row’ refers to the input list of AU Labels, specifically a 12-dimensional binary list; ‘AU list’ is the complete list of individual AUs, containing 12 AU names (e.g., ‘AU1’); ‘Combined AU list’ is the complete list of combined AUs, including 26 AU combination names (e.g., ‘AU6, AU12, AU15, and AU17’), noting that this list is sorted by the number of AUs contained from most to least; ‘dictionary’ is a dictionary comprising the names of 38 generalized AU combinations and their corresponding textual description lists, with each list containing five descriptive texts.

\section{Implementation Details of VQ-AUFace}
\label{sec:impofvqauface}
We developed our methods with the PyTorch framework and employed the Accelerate framework for multi-GPU parallel training. For Stage I training, we used the Adam optimizer to update network parameters. We set the learning rate to 3e-4, chose a batch size of 240, and ran the training for 150,000 steps. During Stage II training, we configured the total number of tokens to 65,536. We set both text and image token sequence lengths to 2,048 and used 8 attention heads. We also utilized T5-large as the text encoder and applied the Adam optimizer again with a learning rate of 3e-4. This stage was performed with a batch size of 16 for 8 epochs. The model was trained on eight GeForce RTX 3090 GPUs, each equipped with 24GB of VRAM, and the entire process took around 48 hours.

\section{Details of the AAAD}
\label{sec:detailsofAAAD}
\subsection{Metrics}
Alignment Accuracy of AU Probability Distributions (AAAD) is a metric used to measure the semantic consistency of facial behaviors between a piece of text and an image. Specifically, the Facial Action Coding System classifies facial behaviors into a series of action units (AUs); when an action unit is activated, we consider that a defined set of facial behaviors has occurred. For images, predicting the activation of specific facial action units from the image yields the occurrence of facial behaviors in the face within the image. For text, predicting the activation of specific facial action units from the text yields the described facial behaviors in the text. When we can predict the probability of activation for specific facial action units from both text and images, we can calculate the similarity between them and use it as a metric for measuring the semantic consistency of facial behaviors between a piece of text and an image.
\subsection{Image AU Recognition}
Facial action unit recognition in facial images is a mature task with numerous methods proposed. We employ the most effective AU recognition method, ME-GraphAU\cite{luo2022learning}, to predict the probability of action unit activation in images. This method achieves the highest recognition accuracy on the BP4D dataset.
\subsection{Text AU Recognition} 
To our knowledge, no researcher has attempted to predict action units from text. This is a multi-classification task where we use T5-Large as the feature extractor and a four-layer perceptron as the classifier, with binary cross-entropy loss as the constraint. The data used for training comes from the BP4D-AUText dataset. We randomly generate five text descriptions for each possible AU combination in the dataset and use these descriptions as training data with the original AU labels as the targets. Ultimately, we obtain a network capable of predicting the probability of action unit activation from any given text. The F1 Score of the used text AU recognition method on the validation set is 0.731.
\subsection{Similarity Calculation} 
Let $P_{i}$ represent the vector of predicted AU activations from the image, and $P_t$ represent the vector of predicted AU activations from the text, both are 12-dimensional vectors where each dimension is a number between 0 and 1, representing the activation probability of the corresponding AU. Specifically, the 12 dimensions correspond to AU1, AU2, AU4, AU6, AU7, AU10, AU12, AU14, AU15, AU17, AU23, and AU24. The cosine similarity between the two activation vectors is calculated as follows:
$$
C = \cos(P_{i}, P_t) = \frac{\sum_{j=1}^{12} P_{i}^{(j)} \cdot P_{t}^{(j)}}{\sqrt{\sum_{j=1}^{12} (P_{i}^{(j)})^2} \cdot \sqrt{\sum_{j=1}^{12} (P_{t}^{(j)})^2}}
$$
where $P_{i}^{(j)}$ and $P_{t}^{(j)}$ represent the $j$-th component of $P_{i}$ and $P_{t}$, respectively.
After obtaining the cosine similarity, it is normalized as follows:
$$
AAAD = \frac{\overline C - C_{min}}{C_{max} - C_{min}}
$$
where $\overline C$ represents the average $C$ of total test set. $C_{max}$ represents the maximal cosine similarity between text-image AU activation vectors calculated from ground truth text-face pairs in the test set, and $C_{min}$ represents the minimum cosine similarity between text-image AU activation vectors calculated from completely shuffled text-face pairs.

This makes AAAD consider the actual distribution of text and image AU activation probabilities in the test set during each calculation, which ensures that the results can be compared with each other even if different text feature extractors are used or evaluated on different test sets.

\section{Implementation Details of Comparison Methods}
\label{app:comparison_setup}

We compare our proposed paradigm with six representative methods from the two existing dominant label-driven paradigms, to verify the superiority of our anatomical description-driven framework.

\textbf{Paradigm 1: Coarse text/attribute label-driven text-to-face methods}: This category includes methods that rely on general facial attribute descriptions or coarse emotion labels for face synthesis.
GLM4\cite{glm2024chatglm} and MidJourney\cite{midjourney_image_generation}: General-purpose large-scale generative models accessed via official online services, with no training on our BP4D-AUText dataset, included for qualitative analysis only.
AnyFace\cite{Sun_Deng_Li_Sun_Ren_Sun}: Classic text-to-face method focusing on static facial attribute synthesis, used in inference-only mode with official pre-trained parameters.
UniPortrait\cite{he2024uniportraitunifiedframeworkidentitypreserving}: State-of-the-art identity-preserving text-to-face method, used in inference-only mode with official pre-trained parameters.
Stable Diffusion 1.5\cite{Rombach_2022_CVPR}: Widely used text-to-image diffusion model, fine-tuned with LoRA on our BP4D-AUText dataset for fair comparison.

\textbf{Paradigm 2: One-hot AU label-driven facial behavior synthesis methods}: This category includes methods that use discrete AU labels as control signals for facial expression synthesis.
GANimation\cite{pumarola2018ganimation}: Classic AU-driven facial animation method, retrained on our BP4D-AUText dataset using AU labels as input, with the same training settings as our baseline model.

All methods are evaluated on the same test set of the BP4D-AUText dataset. GLM4 and MidJourney are only included for qualitative analysis, while the other four methods are evaluated in both quantitative and qualitative analyses.

\section{The Full Input Text in Figure \ref{fig:comparison}}
\label{sec:inputtext}
This section presents the complete text from the leftmost column in Figure \ref{fig:comparison}.

\begin{itemize}
    \item Row 1:\\
    The cheeks are lifted, raising the infraorbital triangle. The lower eyelid is notably raised and straightened, with a bulge indicating tension and a marked narrowing of the eye aperture. The corners of the lips are pulled back and upward, creating an oblique shape to the mouth. The skin below the lower eyelid is pushed up. The upper lip is slightly raised. The overall effect is a complex movement of the facial muscles that results in a significant lifting of the cheeks, lips, and lower lid, with accompanying changes in the skin texture around the eyes and mouth. The central part of the upper lip is pulled upwards vertically while the outer sections are elevated less dramatically, forming an angular contour in the upper lip’s outline. The infraorbital triangle is lifted. Concurrently, the lip corners are slightly lowered, with a sideward tug and a downward angle at the corners, leading to a horizontal stretching of the lips. The chin prominence may appear flattened or develop bulges, and a indentation might occur in the center below the lower lip. An elevation of the upper lip with a more pronounced rise in the center than at the sides straightens the upper lip’s contour. The infraorbital region’s upward movement affects the nasolabial folds. The mouth corners are pulled in tightly, causing the lips to slim at the corners. The skin surrounding the lips is stretched tightly. A deep wrinkle resembling a dimple  appear beyond the mouth corners. The lips’ shape tends towards straight. The skin below the lip corners and on the chin is lifted, contributing to the flattening and stretching of the lower facial skin. 
    \item Row 2:\\
    A notable lifting occurs at the inner and outer regions of the eyebrows, with the inner corners experiencing a minor lift and the outer regions a more accentuated one. The eyebrows are elevated, producing an arched and curved appearance. The eye cover fold is elongated and more noticeable, uncovering the upper eyelid. The center of the upper lip is drawn vertically upwards, with the outer segments of the upper lip also being pulled up, though not as high as the central part. The infraorbital triangle is elevated. The mouth’s corners are notably lifted and slanted obliquely upwards. There is a slightly raised in the infraorbital area.
    \item Row 3:\\
    A subtle uplifting occurs at the inner aspects of the eyebrows, while the skin on the glabella and the forehead above experiences a slight elevation.
    \item Row 4:\\
    The lip corners are extremely tightened, and the skin is pulled inwards around the lip corners. The skin on the chin and lower lip is stretched towards the lip corners, and the lips are stretched and flattened against the teeth. 
    \item Row 5:\\
    As the chin is raised, it produces deep vertical wrinkles below the lower lip. The lips are clenched tightly, nearly hiding the red of the lips. The lower lip is simultaneously elevated and protruded. The chin is elevated. The lips are compressed tightly together, with the lower lip being pushed up and out. There is a noticeable bulging of the skin above the upper lip and below the lower lip. The eyebrows to converge. The inner areas of the eyebrows are slightly lowered on the right and somewhat less so on the left.
    
\end{itemize}

\section{More Qualitative Results}
\label{sec:morequalres}
In this section, we randomly select a face from the dataset and pick some similar descriptions as inputs for face synthesis. The following is the specific result display, where image number 0 is the reference image.

\begin{figure}[h]
    \vspace{5mm}
    \centering
    \includegraphics[width=0.98\linewidth]{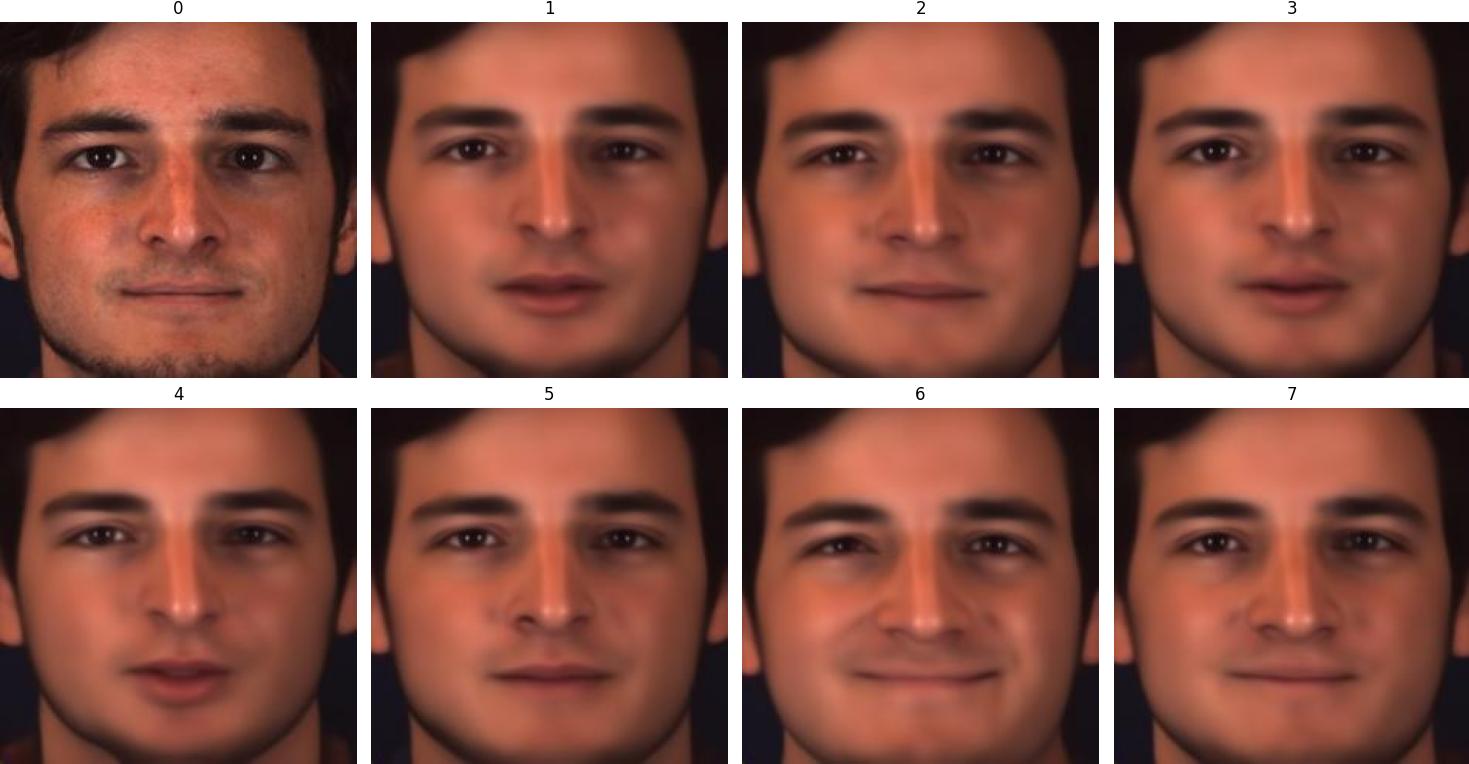}
    \caption{Individual 1}
    \label{fig:cob1}
    \vspace{7mm}
\end{figure}

\begin{figure}[h]
    \centering
    \includegraphics[width=0.98\linewidth]{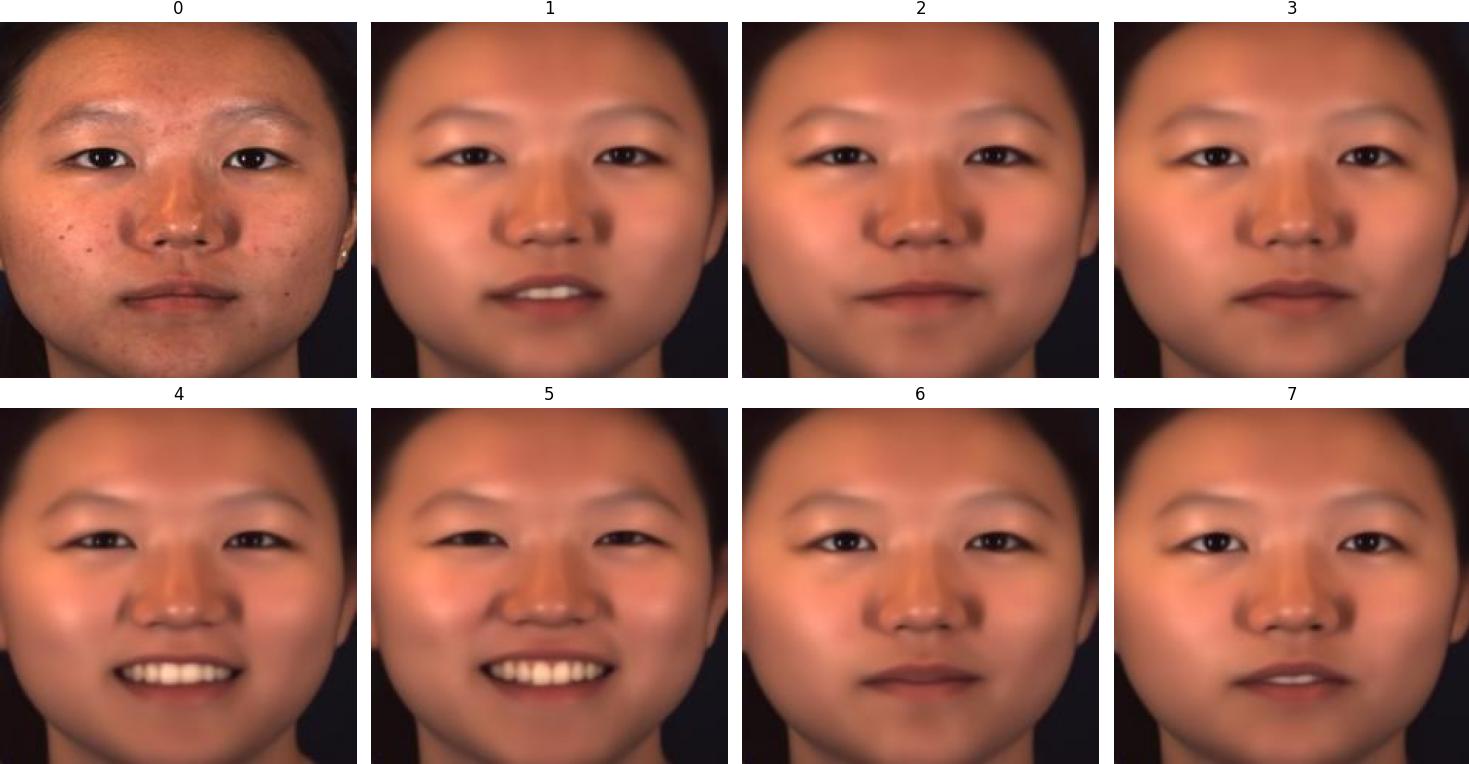}
    \caption{Individual 2}
    \label{fig:cob2}
    \vspace{2.2mm}
\end{figure}

\begin{figure}[h]
\vspace{2.8mm}
    \centering
    \includegraphics[width=0.98\linewidth]{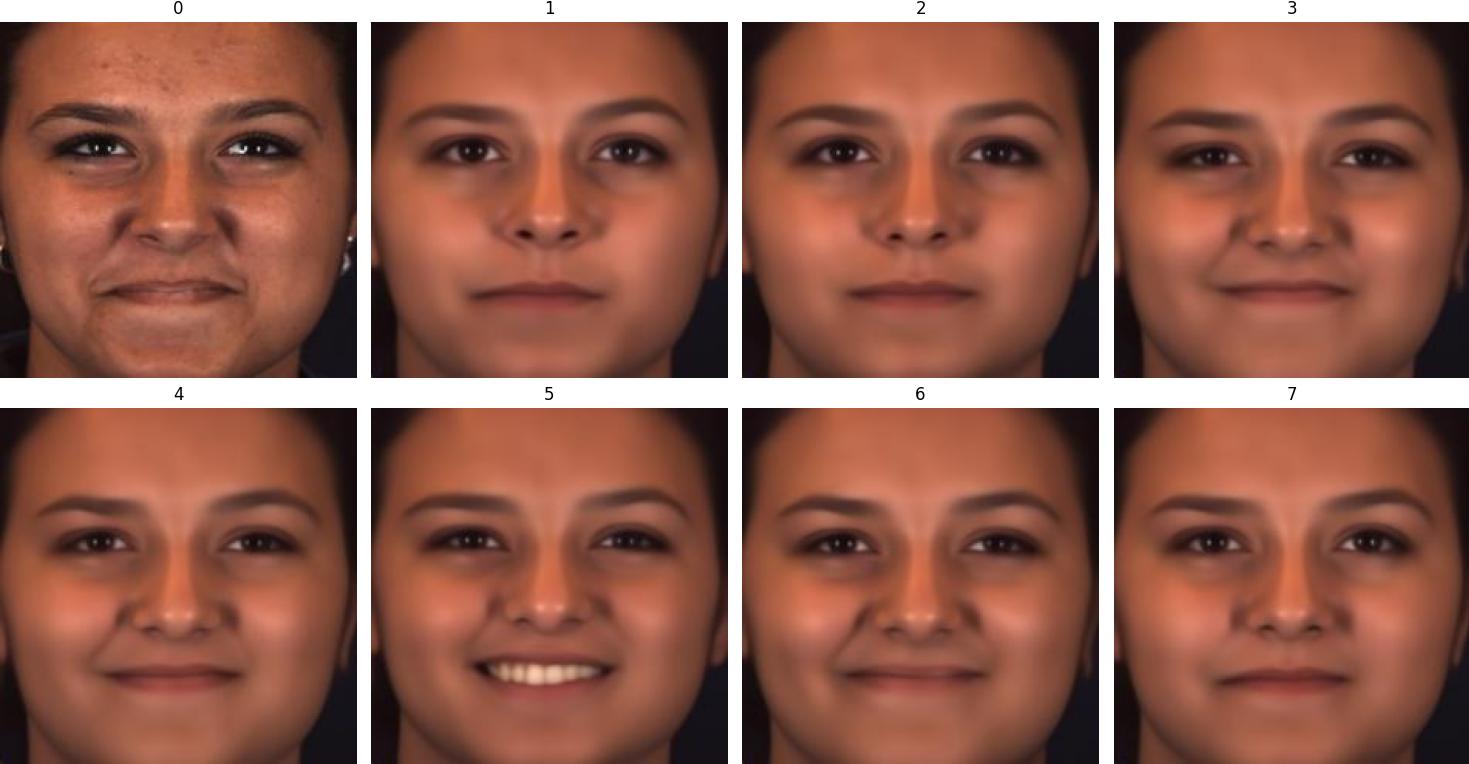}
    \caption{Individual 3}
    \label{fig:cob3}
    \vspace{2.2mm}
\end{figure}

Text Descriptions:
\begin{itemize}
    \item Figure \ref{fig:cob1}: Individual 1
        \begin{enumerate}
            \item A notable raising and slight straightening of the lower eyelid occur, causing a slight bulge, and the eye’s opening shows a marked to pronounced reduction in width.
            \item The inner and outer sections of the eyebrows are visibly lifted, with the inner corners slightly elevated and the outer sections more distinctly lifted. The eyebrows move upwards, forming an arched and curved shape. The fold of the eye cover is extended, making it more noticeable and exposing the upper eyelid. A direct upward movement of the upper lip’s center, contrasted with a milder lift at the sides, results in an angular modification to the upper lip’s contour. The infraorbital triangle experiences an upward force. The corners of the lips are concurrently drawn slightly down and outward, leading to a lateral stretching of the lips. The chin’s boss  seem to flatten or have bulges, and a central indentation might appear below the lower lip. A minimal descent of the mouth corners is accompanied by the lips’ tightening and inward movement, concealing the red part of the lips. There is a pronounced lift and slight flattening of the lower eyelid, leading to a slight bulge, and the eye’s aperture exhibits a clear to strong narrowing.
            \item A slight downturn occurs at the corners of the mouth, while the lips are tensed and pulled towards the center, diminishing the red part of the lips’ visibility. A pronounced lift is seen in the outer aspect of the eyebrows. The exposure of the eyelid fold and the skin is more pronounced. The lower eyelid is raised markedly and straightened slightly, causing slight bulging, and the narrowing of the eye aperture is marked to pronounced.
            \item The lower eyelid is significantly elevated and slightly straightened, resulting in a slight protrusion, and the eye opening is narrowed to a marked or pronounced degree. The lip corners are pulled down slightly, with some lateral pulling and angling down of the corners.
            \item Raising the lower eyelid causes it to become taut and form a bulge, thereby constricting the eye’s aperture. The corners of the lips are visibly lifted and directed upwards at an angle, deepening the nasolabial furrow. A slight elevation of the infraorbital triangle is observed. There is a marked tightening of the lip corners, accompanied by wrinkling of the skin as it is pulled inwardly around the lips’ edges. The skin on the chin and lower lip is stretched in the direction of the lip corners, and the lips are elongated and pressed flat against the teeth.
            \item Lifting the cheeks causes the infraorbital triangle to rise. The lower eyelid is significantly lifted and tensed, with a bulge that points to tension and a notable decrease in the width of the eye opening. The lips’ corners are retracted and elevated, forming an oblique mouth contour. The skin under the lower eyelid is pushed up. The upper lip is slightly elevated as well. The intricate play of facial muscles results in a marked lifting of the cheeks, lips, and lower eyelid, along with associated changes in the skin’s texture around the eyes and mouth. The central part of the upper lip is lifted more than the sides, straightening the upper lip line. The infraorbital area’s upward push affects the nasolabial folds. The mouth corners are drawn tightly inwards, making the lips appear narrower at the corners. The skin around the lips is pulled taught. A significant dimple-like wrinkle could extend from the mouth corners. The lips assume a straighter form. The skin under the lip corners and the chin is elevated, causing the lower facial skin to flatten and stretch.
            \item The cheeks are lifted, raising the infraorbital triangle. The lower eyelid is notably raised and straightened, with a bulge indicating tension and a marked narrowing of the eye aperture. The corners of the lips are pulled back and upward, creating an oblique shape to the mouth. The skin below the lower eyelid is pushed up. The upper lip is slightly raised. The overall effect is a complex movement of the facial muscles that results in a significant lifting of the cheeks, lips, and lower lid, with accompanying changes in the skin texture around the eyes and mouth. A distinct tensing of the facial muscles causes the lips to tighten considerably, with the mouth corners drawn snugly towards the center. The lips become slimmed down, and the skin on the chin and lower lip is pulled towards the mouth corners. The chin prominence seems flattened and elongated, and the nasolabial folds are more pronounced. A minimal descent of the mouth corners is accompanied by the lips’ tightening and inward movement, concealing the red part of the lips.
        \end{enumerate}
        \item Figure \ref{fig:cob2} Individual 2
        \begin{enumerate}
            \item There is a pronounced lift and slight flattening of the lower eyelid, leading to a slight bulge, and the eye’s aperture exhibits a clear to strong narrowing. The center of upper lip is drawn straight up, the outer portions of upper lip are drawn up but not as high as the center. The infraorbital triangle is pushed up.
            \item The upper lip is elevated with the center portion raised more significantly than the sides, resulting in a straightening of the upper lip line. The infraorbital area is pushed upward. The corners of the mouth are drawn tightly inward, causing the lips to narrow at the corners. The skin around the lips is pulled taut. A deep dimple-like wrinkle  extend beyond the corners of the mouth. The overall shape of the lips becomes more straight than curved. The skin below the lip corners and the chin area is pulled up, contributing to a flattening and stretching of the lower facial skin. A marked constriction of the lips is evident in the facial expression, with the mouth corners being drawn tightly inwards. The lips are narrowed, and the skin on the lower lip and chin moves towards the mouth corners. The chin’s projection appears flattened and extended, and the nasolabial grooves are more accentuated. A slight elevation is noticeable at the inner corners of the eyebrows, with a corresponding slight lifting of the skin on the glabella and the forehead above. There is a pronounced lift and slight flattening of the lower eyelid, leading to a slight bulge, and the eye’s aperture exhibits a clear to strong narrowing. The lips are severely pressed together, with bulging skin above and below the red parts, and narrowing of the lips.
            \item A distinct tensing of the facial muscles causes the lips to tighten considerably, with the mouth corners drawn snugly towards the center. The lips become slimmed down, and the skin on the chin and lower lip is pulled towards the mouth corners. The chin prominence seems flattened and elongated, and the nasolabial folds are more pronounced. The lower eyelid is significantly lifted and slightly aligned, causing a slight swelling, and the eye’s aperture narrowing is evident to pronounced.
            \item The infraorbital triangle is elevated as the cheeks are lifted. The lower eyelid is significantly lifted and flattened, showing a bulge that signifies tension and a clear reduction in the size of the eye opening. The mouth’s corners are retracted and pulled skyward, forming an oblique mouth shape. The skin under the lower eyelid is pushed upward. The upper lip experiences a slight elevation. The collective motion of the facial muscles produces a pronounced lifting of the cheeks, lips, and lower eyelid, along with visible changes in the skin’s texture around the orbital and oral areas. The center of upper lip is drawn straight up, the outer portions of upper lip are drawn up but not as high as the center. The infraorbital triangle is pushed up.
            \item The cheeks are lifted, raising the infraorbital triangle. The lower eyelid is notably raised and straightened, with a bulge indicating tension and a marked narrowing of the eye aperture. The corners of the lips are pulled back and upward, creating an oblique shape to the mouth. The skin below the lower eyelid is pushed up. The upper lip is slightly raised. The overall effect is a complex movement of the facial muscles that results in a significant lifting of the cheeks, lips, and lower lid, with accompanying changes in the skin texture around the eyes and mouth. An elevation of the upper lip with a more pronounced rise in the center than at the sides straightens the upper lip’s contour. The infraorbital region’s upward movement affects the nasolabial folds. The mouth corners are pulled in tightly, causing the lips to slim at the corners. The skin surrounding the lips is stretched tightly. A deep wrinkle resembling a dimple  appear beyond the mouth corners. The lips’ shape tends towards straight. The skin below the lip corners and on the chin is lifted, contributing to the flattening and stretching of the lower facial skin.
            \item Lift your cheeks without activately raising up the lip corners. The infraorbital triangle is raised slightly. There is a pronounced lift and slight flattening of the lower eyelid, leading to a slight bulge, and the eye’s aperture exhibits a clear to strong narrowing. The center of the upper lip is drawn vertically upwards, with the outer segments of the upper lip also being pulled up, though not as high as the central part. The infraorbital triangle is elevated. The lips are maximally tensed and the red parts are maximally narrowed, leading to intense wrinkling and bulging around the margins of the red parts of both lips.
            \item The lower eyelid is significantly elevated and slightly straightened, resulting in a slight protrusion, and the eye opening is narrowed to a marked or pronounced degree. The midpoint of the upper lip is pulled directly upwards, but the outer regions of the upper lip are drawn up less than the midpoint. The infraorbital triangle is thrust upwards. The lips are tensed to the utmost extent and the red portions are minimized to the maximum, resulting in severe wrinkling and protrusion around the edges of the red sections of both lips.
        \end{enumerate}
        \item Figure \ref{fig:cob3} Individual 3
        \begin{enumerate}
            \item Drawing the mouth corners back and up leads to a tightening and straightening of the lip line. The lower lip is stretched and pressed against the upper lip. The chin and the region of the lower lip experience an upward push, causing the chin’s skin to wrinkle and the lower lip to be raised, narrowing the red part of the lips further. This action induces a depression to form under the lower lip’s center. The lips appearance is tight and compressed, with a clear transformation of the lips’ normal contour. The facial muscles show a pronounced tightening that affects the lips, pulling the corners of the mouth tightly to the inside. The lips become narrow, and the skin on the chin and lower lip stretches towards the corners of the mouth. The chin’s roundness is flattened and elongated, and the nasolabial folds are deepened. A lifting of the chin area generates deep vertical folds beneath the lower lip. The lips are tightly retracted, with the red lip area almost concealed. The lower lip experiences an outward push as it is raised. The chin is elevated. The lips are compressed tightly together, with the lower lip being pushed up and out. There is a noticeable bulging of the skin above the upper lip and below the lower lip.
            \item Lift your cheeks without activately raising up the lip corners. The infraorbital triangle is raised slightly. There is a pronounced lift and slight flattening of the lower eyelid, leading to a slight bulge, and the eye’s aperture exhibits a clear to strong narrowing. The lips are tightened maximally and the red parts are narrowed maximally, creating extreme wrinkling and bulging around the margins of the red parts of both lips. A severe compression of the lips occurs, causing an outward pushing of the skin above and below the red parts, and a tightening of the lips.
            \item The infraorbital triangle is elevated as the cheeks are lifted. The lower eyelid is significantly lifted and flattened, showing a bulge that signifies tension and a clear reduction in the size of the eye opening. The mouth’s corners are retracted and pulled skyward, forming an oblique mouth shape. The skin under the lower eyelid is pushed upward. The upper lip experiences a slight elevation. The collective motion of the facial muscles produces a pronounced lifting of the cheeks, lips, and lower eyelid, along with visible changes in the skin’s texture around the orbital and oral areas. The lips are tensed to the utmost extent and the red portions are minimized to the maximum, resulting in severe wrinkling and protrusion around the edges of the red sections of both lips. There is a severe pressing of the lips together, resulting in a bulging of the skin above and below the red portions, with a constriction of the lips.
            \item With the cheeks raised, the infraorbital triangle is lifted. The lower eyelid is notably lifted and made taut, with a bulge indicating strain and a significant constriction of the eye’s aperture. The corners of the lips are pulled backward and upward, resulting in an angled mouth shape. The skin below the lower eyelid is forced upward. The upper lip sees a slight rise. The overall movement of the facial muscles creates a notable lifting effect on the cheeks, lips, and lower eyelid, with concurrent changes in the skin texture near the eyes and mouth. The middle of the upper lip is lifted straight up, whereas the lateral parts of the upper lip are raised yet not as much as the middle. The infraorbital area is pushed up. The lips are tightened maximally and the red parts are narrowed maximally, creating extreme wrinkling and bulging around the margins of the red parts of both lips. The lips are severely pressed together, with bulging skin above and below the red parts, and narrowing of the lips.
            \item The cheeks are lifted, raising the infraorbital triangle. The lower eyelid is notably raised and straightened, with a bulge indicating tension and a marked narrowing of the eye aperture. The corners of the lips are pulled back and upward, creating an oblique shape to the mouth. The skin below the lower eyelid is pushed up. The upper lip is slightly raised. The overall effect is a complex movement of the facial muscles that results in a significant lifting of the cheeks, lips, and lower lid, with accompanying changes in the skin texture around the eyes and mouth. The middle of the upper lip is lifted straight up, whereas the lateral parts of the upper lip are raised yet not as much as the middle. The infraorbital area is pushed up. The lips are maximally constricted and the red areas are maximally constricted, causing extreme wrinkling and bulging along the perimeters of the red parts of both lips.
            \item With the cheeks raised, the infraorbital triangle is lifted. The lower eyelid is notably lifted and made taut, with a bulge indicating strain and a significant constriction of the eye’s aperture. The corners of the lips are pulled backward and upward, resulting in an angled mouth shape. The skin below the lower eyelid is forced upward. The upper lip sees a slight rise. The overall movement of the facial muscles creates a notable lifting effect on the cheeks, lips, and lower eyelid, with concurrent changes in the skin texture near the eyes and mouth. An upward pull on the middle of the upper lip, with a less obvious lift on the sides, forms an angular shape in the upper lip. The area of the infraorbital triangle is raised. The corners of the lips are gently drawn downwards, with a sideways and downward tension that stretches the lips laterally. The chin’s fullness  seem to flatten or present with bulges, and a central dip might form beneath the lower lip. The center of the upper lip rises more prominently than its edges, causing the upper lip line to become straight. The infraorbital region is lifted. The mouth corners are pulled tightly towards the center, narrowing the lips at their edges. The skin around the lips is stretched. A deep dimple-like wrinkle might extend past the mouth corners. The lips’ general shape becomes more linear than curved. The skin beneath the lip corners and the chin is lifted, leading to a flattening and extension of the lower facial skin.
            \item The upper lip is raised, with the middle section showing a greater increase in height than the side sections, leading to a straightening effect on the upper lip’s line. The infraorbital area experiences an upward force. The mouth corners are pulled snugly towards each other, narrowing the lips at the corners. The skin around the lips is pulled taut. A pronounced dimple-like wrinkle might extend past the mouth corners. The overall shape of the lips is more straight than curved. The skin under the lip corners as well as the chin area is lifted, resulting in a flattening and stretching of the skin on the lower face. The mouth corners are raised and extended laterally, forming a more pointed angle. The muscles that frame the mouth are stiff, slightly narrowing the mouth’s opening. The skin around the nose and mouth is drawn snugly, resulting in lines that stretch from the nose to the mouth’s corners. A slight bulge is visible on the lower lip. A marked constriction of the lips is evident in the facial expression, with the mouth corners being drawn tightly inwards. The lips are narrowed, and the skin on the lower lip and chin moves towards the mouth corners. The chin’s projection appears flattened and extended, and the nasolabial grooves are more accentuated
        \end{enumerate}
\end{itemize}

\section{Full AU Text Descriptions}
\label{sec:fullautext}
\subsection{Individual AU}
\begin{itemize}
    \item AU1: 
    \begin{enumerate}
        \item The inner corners of the eyebrows are lifted slightly, the skin of the glabella and forehead above it is lifted slightly and wrinkles deepen slightly and a trace of new ones form in the center of the forehead.
        \item A subtle uplifting occurs at the inner aspects of the eyebrows, while the skin on the glabella and the forehead above experiences a slight elevation, leading to a slight intensification of wrinkles and the formation of new ones in the middle of the forehead.
        \item There is a slight raising of the innermost parts of the eyebrows, and a corresponding slight lifting of the skin over the glabella and forehead, causing the wrinkles to become slightly more pronounced and initiating the appearance of new lines in the central forehead.
        \item Slightly lifting the inner parts of the eyebrows results in a concurrent slight elevation of the skin on the glabella and the forehead above, slightly deepening the existing wrinkles and causing new ones to start forming in the middle of the forehead.
        \item A slight elevation is noticeable at the inner corners of the eyebrows, with a corresponding slight lifting of the skin on the glabella and the forehead above, which slightly deepens the wrinkles and allows new ones to begin to appear in the center of the forehead.

    \end{enumerate}
        \item AU2: 
    \begin{enumerate}
        \item The outer part of the eyebrow raise is pronounced. The wrinkling above the right outer eyebrow has increased markedly, and the wrinkling on the left is pronounced. Increased exposure of the eye cover fold and skin is pronounced.
        \item The lateral region of the eyebrows exhibits a significant lift. The creasing above the right lateral eyebrow has noticeably intensified, and the creases above the left lateral eyebrow are pronounced. There is an increased visibility of the eyelid fold and the skin appears more pronounced.
        \item A notable elevation is observed in the outer section of the eyebrows. The wrinkling above the outer right eyebrow has significantly augmented, and the wrinkling above the left outer eyebrow is quite pronounced. The exposure of the eyelid covering fold and the skin is clearly enhanced.
        \item The outer portion of the eyebrows shows a pronounced raise. There is a marked increase in wrinkling above the outer right eyebrow, and the wrinkling above the left outer eyebrow is also pronounced. There is an increased prominence in the visibility of the eyelid fold and the skin.
        \item A pronounced lift is seen in the outer aspect of the eyebrows. The wrinkling above the right outer eyebrow has notably increased, and the wrinkling above the left is also pronounced. The exposure of the eyelid fold and the skin is more pronounced.

    \end{enumerate}
    \item AU4: 
    \begin{enumerate}
        \item Vertical wrinkles appear in the glabella and the eyebrows are pulled together. The inner parts of the eye-brows are pulled down a trace on the right and slightly on the left with traces of wrinkling at the corners.
        \item Vertical creases emerge in the glabella, and the eyebrows are drawn towards each other. The inner sections of the eyebrows are slightly descended on the right and to a lesser extent on the left, with faint signs of wrinkling at the edges.
        \item The glabella develops vertical furrows, and the eyebrows come closer together. There is a slight pulling down of the inner parts of the eyebrows on the right and a bit less so on the left, accompanied by slight wrinkling at the corners.
        \item Vertical lines appear in the glabella, causing the eyebrows to converge. The inner areas of the eyebrows are slightly lowered on the right and somewhat less so on the left, with subtle wrinkling evident at the corners.
        \item The glabella shows vertical wrinkles, and the eyebrows are brought together. A slight downward pull is noticed in the inner parts of the eyebrows, more so on the right and slightly less on the left, with trace amounts of wrinkling at the corners.

    \end{enumerate}
    \item AU6: 
    \begin{enumerate}
        \item lift your cheeks without activately raising up the lip corners. The infraorbital furrow has deepened slightly and bags or wrinkles under the eyes must in crease. The infraorbital triangle is raised slightly.
        \item Elevate your cheeks without actively lifting the corners of the lips. The infraorbital crease has become slightly more pronounced, and bags or wrinkles beneath the eyes should increase. The infraorbital triangle is slightly elevated.
        \item Raise your cheekbones without consciously lifting the mouth’s corners. The infraorbital groove has deepened marginally, and puffiness or creases under the eyes need to expand. The infraorbital area is slightly lifted.
        \item Enhance the cheeks without deliberately raising the lip corners. The infraorbital fold has moderately deepened, and bags or lines under the eyes should grow. The infraorbital region is slightly raised.
        \item Boost your cheeks without actively elevating the lips’ edges. The infraorbital depression has subtly intensified, and pouches or wrinkles under the eyes must increase. The infraorbital area is slightly uplifted.

    \end{enumerate}
    \item AU7: 
    \begin{enumerate}
        \item The lower eyelid is raised markedly and straightened slightly, causing slight bulging, and the narrowing of the eye aperture is marked to pronounced.
        \item The lower eyelid is significantly elevated and slightly straightened, resulting in a slight protrusion, and the eye opening is narrowed to a marked or pronounced degree.
        \item There is a pronounced lift and slight flattening of the lower eyelid, leading to a slight bulge, and the eye’s aperture exhibits a clear to strong narrowing.
        \item A notable raising and slight straightening of the lower eyelid occur, causing a slight bulge, and the eye’s opening shows a marked to pronounced reduction in width.
        \item The lower eyelid is significantly lifted and slightly aligned, causing a slight swelling, and the eye’s aperture narrowing is evident to pronounced.

    \end{enumerate}
    \item AU10: 
    \begin{enumerate}
        \item The center of upper lip is drawn straight up, the outer portions of upper lip are drawn up but not as high as the center. The infraorbital triangle is pushed up, the nasolabial furrow is deepened.
        \item The central part of the upper lip is pulled directly upwards, while the outer sections of the upper lip are also elevated but not to the same extent as the center. The infraorbital triangle is shifted upwards, and the nasolabial crease is intensified.
        \item The middle of the upper lip is lifted straight up, whereas the lateral parts of the upper lip are raised yet not as much as the middle. The infraorbital area is pushed up, and the nasolabial groove is deepened.
        \item The center of the upper lip is drawn vertically upwards, with the outer segments of the upper lip also being pulled up, though not as high as the central part. The infraorbital triangle is elevated, and the nasolabial furrow is made deeper.
        \item The midpoint of the upper lip is pulled directly upwards, but the outer regions of the upper lip are drawn up less than the midpoint. The infraorbital triangle is thrust upwards, and the nasolabial fold is deepened.

    \end{enumerate}
    \item AU12: 
    \begin{enumerate}
        \item The corners of the lips are markedly raised and angled up obliquely. The nasolabial furrow has deepened slightly and is raised obliquely slightly. The infraorbital triangle is raised slightly.
        \item The corners of the mouth are significantly elevated and tilted upwards at an angle. The nasolabial crease has moderately deepened and is slightly raised at an angle. The infraorbital triangle experiences a slight uplift.
        \item The mouth’s corners are notably lifted and slanted obliquely upwards. The nasolabial furrow has slightly intensified and is slightly elevated at an angle. There is a slightly raised in the infraorbital area.
        \item A pronounced lifting and diagonal angling upwards occur at the corners of the lips. The nasolabial fold has slightly deepened and is slightly raised in an oblique direction. The infraorbital triangle is slightly raised.
        \item The corners of the lips are markedly raised and directed upwards obliquely. The nasolabial furrow has slightly deepened and is also slightly raised in an oblique fashion. A slight elevation is seen in the infraorbital triangle.

    \end{enumerate}
    \item AU14: 
    \begin{enumerate}
        \item The lip corners are extremely tightened, and the wrinkling as skin is pulled inwards around the lip corners is severe. The skin on the chin and lower lip is stretched towards the lip corners, and the lips are stretched and flattened against the teeth.
        \item The lip corners are excessively constricted, and the wrinkling of the skin as it is drawn inward around the lip edges is pronounced. The skin on the chin and lower lip is pulled toward the lip corners, and the lips are stretched and smoothed against the teeth.
        \item The corners of the lips are tightly drawn, with the resulting wrinkling of the skin inward around the lips being intense. The skin on the chin and lower lip is drawn towards the lips’ corners, and the lips are extended and flattened against the teeth.
        \item There is a marked tightening of the lip corners, accompanied by severe wrinkling of the skin as it is pulled inwardly around the lips’ edges. The skin on the chin and lower lip is stretched in the direction of the lip corners, and the lips are elongated and pressed flat against the teeth.
        \item The lip corners are very tightly knit, leading to significant wrinkling of the skin as it gets pulled inward around the lip area. The skin on the chin and lower lip is directed towards the lip corners, and the lips are stretched and made to lie flat against the teeth.

    \end{enumerate}
    \item AU15: 
    \begin{enumerate}
        \item The lip corners are pulled down, with some lateral pulling and angling down of the corners, and slight bulges and wrinkles appear beyond the lip corners.
        \item The corners of the lips are lowered, experiencing some outward pulling and a downward angling, and small bulges and wrinkles become visible outside the lip corners.
        \item There is a downward traction on the lip corners, with a side pulling and an angling down movement of the corners, and protrusions and creases emerge past the lip corners.
        \item A descent of the lip corners is noted, along with a bit of lateral tension and a downward slanting of the corners, and minor bulges and wrinkles become evident beyond the lip corners.
        \item Downward pulling is applied to the lip corners, accompanied by some lateral pulling and a downturn of the corners, and small bulges and wrinkles are formed outside the lip corners.

    \end{enumerate}
    \item AU17: 
    \begin{enumerate}
        \item The chin boss shows severe to extreme wrinkling as it is pushed up severely, and the lower lip is pushed up and out markedly.
        \item The chin prominence exhibits intense to extreme wrinkling as it is significantly elevated, and the lower lip is notably raised and protruded.
        \item The chin boss demonstrates severe to extreme creasing as it experiences a marked upward push, while the lower lip is prominently lifted and thrust out.
        \item The chin protuberance shows intense to extreme wrinkling due to a substantial upward force, and the lower lip is significantly elevated and extruded.
        \item Severe to extreme wrinkling is seen on the chin boss as it is pushed up substantially, and the lower lip is notably raised and projected outward.

    \end{enumerate}
    \item AU23: 
    \begin{enumerate}
        \item The lips are tightened maximally and the red parts are narrowed maximally, creating extreme wrinkling and bulging around the margins of the red parts of both lips.
        \item The lips are tensed to the utmost extent and the red portions are minimized to the maximum, resulting in severe wrinkling and protrusion around the edges of the red sections of both lips.
        \item The lips are maximally constricted and the red areas are maximally constricted, causing extreme wrinkling and bulging along the perimeters of the red parts of both lips.
        \item The lips are tightened to the fullest extent and the red sections are reduced to the fullest extent, generating profound wrinkling and bulging around the borders of the red parts of both lips.
        \item The lips are maximally tensed and the red parts are maximally narrowed, leading to intense wrinkling and bulging around the margins of the red parts of both lips.

    \end{enumerate}
    \item AU24: 
    \begin{enumerate}
        \item The lips are severely pressed together, severely bulging skin above and below the red parts, with severe narrowing of the lips and wrinkling above the upper lip.
        \item The lips are tightly compressed against each other, causing a severe protrusion of the skin both above and below the colored sections, accompanied by a severe reduction in lip width and significant wrinkling above the upper lip.
        \item There is a severe pressing of the lips together, resulting in a severe bulging of the skin above and below the red portions, with a severe constriction of the lips and pronounced wrinkling over the upper lip.
        \item The lips are forcefully pressed together, leading to a severe bulge in the skin above and below the red areas, along with a severe narrowing of the lips and severe wrinkling above the upper lip.
        \item A severe compression of the lips occurs, causing a severe outward pushing of the skin above and below the red parts, and a severe tightening of the lips with severe wrinkling above the upper lip.

    \end{enumerate}
    
\end{itemize}

\subsection{Combined AU}
\begin{itemize}
    \item AU6, AU12, AU15 and AU17:
        \begin{enumerate}
        \item The eyelids are raised, with the skin around the eyes becoming more tense and stretched, particularly in the outer corners where crow’s feet wrinkles  appear. The inner corners of the eyebrows are pulled upwards and inwards, while the outer edges of the eyebrows are raised and drawn together, causing a vertical furrow between the eyebrows. The upper eyelids are retracted, making the eyes appear more open and exposed. The cheeks are raised, causing the skin to bunch up and create a series of horizontal wrinkles extending from the outer corners of the eyes towards the temples. The lips are parted, with the upper lip pulled up and back, potentially exposing the upper teeth, while the lower lip  move slightly downwards and outwards, though not as much as the upper lip. The chin is lifted, resulting in a tensing and wrinkling of the skin below the lower lip, creating a dimple-like appearance directly under the center of the chin.
	\item The raising of the eyelids leads to a tightening of the skin around the orbital area, especially at the outer corners where crow’s feet  start to show. The inner parts of the eyebrows are drawn upward and inward, while the outer parts elevate and converge, forming a vertical crease between the brows. The upper eyelids pull back, giving the eyes a wider and more uncovered look. The cheeks are elevated, causing the flesh to gather and form horizontal lines from the eye’s outer corners to the temples. The mouth is slightly open, with the top lip raised and pulled backward, possibly revealing the upper teeth, and the bottom lip slightly descending and protruding, but less so than the top lip. The chin is elevated, causing the skin beneath the lower lip to tense and wrinkle, creating a dimple-like effect beneath the chin’s center.
	\item As the eyelids lift, the skin surrounding the eyes becomes taut and stretched, with a pronounced effect at the outer corners where crow’s feet might emerge. The inner ends of the eyebrows are lifted toward the center, and the outer ends rise and come together, leading to a vertical line between the eyebrows. The upper eyelids are drawn back, enhancing the openness of the eyes. The cheeks are lifted, resulting in a gathering of skin that forms horizontal creases from the eyes’ outer edges to the temple area. The lips are separated, with the upper lip raised and retracted, possibly uncovering the upper teeth, while the lower lip slightly descends and protrudes less than the upper lip. The chin is raised, causing the skin under the lower lip to pucker and form a dimple-like indentation beneath the chin’s midpoint.
	\item Elevating the eyelids causes the skin around them to tense, especially at the outer eye area where crow’s feet can form. The eyebrows’ inner sections are pulled up and inward, while the outer sections elevate and knit together, forming a vertical wrinkle between the brows. The upper eyelids are pulled back, making the eyes seem more exposed. The cheeks are lifted, leading to a bunching of skin that creates horizontal lines from the eyes’ outer corners to the temple region. The lips are slightly apart, with the upper lip raised and drawn back, which  reveal the upper teeth, and the lower lip exhibiting a lesser degree of downward and outward movement. The chin is lifted, causing the skin below the lower lip to wrinkle and form a dimple-like appearance directly beneath the chin’s center.
	\item The action of lifting the eyelids tightens the skin around the eyes, with a notable effect at the outer corners where crow’s feet could become visible. The inner portions of the eyebrows move upward and inward, while the outer parts rise and come together, creating a vertical crease between the eyebrows. The upper eyelids are retracted, giving the eyes a more open appearance. The cheeks are raised, resulting in skin folding that generates horizontal wrinkles from the outer edges of the eyes to the temple areas. The lips are opened slightly, with the upper lip pulled up and backward, possibly exposing the upper teeth, and the lower lip showing a less pronounced movement downward and outward. The chin is elevated, leading to a tensing and creasing of the skin under the lower lip, forming a dimple-like feature directly under the chin’s central point.
    \end{enumerate}
    
    \item AU6, AU12, AU17 and AU23:
        \begin{enumerate}
        \item The cheeks are raised, causing the skin below the eyes to smooth out and the area beneath the lower eyelids to become more prominent. The inner corners of the eyebrows are pulled slightly upward, while the outer corners remain relatively unchanged, leading to a subtle change in the shape of the eyebrows. The corners of the mouth are lifted and drawn back, creating a stretching of the skin around the mouth. This action results in the appearance of vertical wrinkles above the upper lip and horizontal wrinkles below the lower lip. The lower lip is raised, causing it to protrude slightly, and the upper lip  also be elevated, although less so than the lower lip. This can lead to a more open exposure of the teeth and gums. The chin is pushed upward, leading to the formation of deeper horizontal wrinkles across the chin area. A narrowing of the lips and a tightening of the facial features around the mouth, with the lower face appearing more tense and the expression being one of surprise or startle.
	\item Elevating the cheeks results in the skin under the eyes becoming smoother and the region under the lower eyelids becoming more accentuated. The inner ends of the eyebrows are slightly raised, with the outer ends staying relatively the same, which slightly alters the form of the eyebrows. The mouth’s corners are hoisted and retracted, causing the skin around the mouth to stretch. This movement gives rise to vertical lines above the upper lip and horizontal creases below the lower lip. The lower lip is lifted, making it slightly more prominent, and the upper lip is also raised, albeit to a lesser degree. This can expose more of the teeth and gums. The chin is elevated, contributing to the creation of more pronounced horizontal wrinkles across the chin. The lips narrow and the facial muscles around the mouth tighten, making the lower part of the face seem more strained and conveying a look of surprise or astonishment.
	\item As the cheeks are lifted, the skin under the orbital area smoothes and the tissue below the lower eyelids becomes more noticeable. The inner parts of the eyebrows experience a slight upward pull, while the outer parts remain largely static, subtly modifying the eyebrow shape. The mouth’s corners are pulled up and backward, stretching the skin encircling the mouth and resulting in vertical creases over the upper lip and horizontal lines beneath the lower lip. The lower lip is elevated, causing it to stick out a bit, and the upper lip also rises, but not as much as the lower. This can lead to a greater visibility of the teeth and gums. The chin is thrust up, causing deeper horizontal creases to form on the chin. The lips become narrower, and the muscles around the mouth tense, giving the lower face a tenser appearance and an expression of being startled or surprised.
	\item With the cheeks lifted, the area beneath the eyes becomes smoother and the tissue under the lower eyelids is highlighted. The inner portions of the eyebrows are slightly raised, while the outer parts are mostly unchanged, slightly changing the eyebrow’s contour. The corners of the mouth are raised and pulled toward the back, which stretches the skin around the mouth and forms vertical lines above the upper lip and horizontal lines below the lower lip. The lower lip is raised, making it a bit more projecting, and the upper lip also rises, though less than the lower lip, potentially revealing more of the dental and gum area. The chin is pushed up, causing more pronounced horizontal creases to appear on the chin. The lips narrow, and the area around the mouth tightens, making the lower face seem more tense and expressing a look of surprise or shock.
	\item When the cheeks are elevated, the skin under the eyes smoothens and the area under the lower eyelids stands out more. The inner sections of the eyebrows are drawn slightly upward, with the outer sections remaining almost the same, thus slightly altering the eyebrow’s form. The mouth’s corners are lifted and pulled back, leading to the stretching of the skin around the mouth and the creation of vertical wrinkles above the upper lip and horizontal wrinkles below the lower lip. The lower lip is lifted, causing it to bulge slightly, while the upper lip also rises, but to a lesser extent. This  expose the teeth and gums more. The chin is moved up, resulting in deeper horizontal wrinkles across the chin. The lips become narrower, and the muscles around the mouth contract, giving the lower face a tenser look and an expression of being startled or surprised.
    \end{enumerate}
        
    \item AU1, AU2 and AU4:
        \begin{enumerate}
        \item A lifting of both the inner and outer corners of the eyebrows, with the skin between the brows and on the forehead becoming slightly more tense.  This tension results in a deepening of existing wrinkles and the formation of new, subtle wrinkles across the center of the forehead.  The overall shape of the eyebrows  become slightly more arched, with the outer halves elevated and the inner halves pulled upwards, creating a broader opening of the eyes.  The forehead displays horizontal lines, and there is a noticeable bunching of the skin in the central area, potentially forming a small peak or ridge.  An oblique line or bunching of the muscle can be seen extending from the inner brow corner upwards and outwards towards the middle of the forehead.
	\item Both the inner and outer ends of the eyebrows are elevated, causing the skin between the eyebrows and on the forehead to tense up slightly. This increased tension contributes to the intensification of pre-existing wrinkles and the emergence of new, fine wrinkles across the forehead’s middle. The eyebrows  take on a slightly more curved shape, with the outer segments raised and the inner segments drawn up, which enlarges the visible area of the eyes. Horizontal creases are visible on the forehead, and there might be a discernible gathering of skin in the central region, possibly forming a slight elevation or ridge. A diagonal line or muscle contraction is observable starting from the inner brow corner and extending upward and outward toward the forehead’s center.
	\item The eyebrows are lifted at both their inner and outer corners, leading to a slight tensing of the skin between the brows and on the forehead. This tensing causes a deepening of current wrinkles and the appearance of new, faint wrinkles across the forehead’s central zone. The eyebrows might appear more arched, with the outer parts raised and the inner parts pulled up, which opens up the eyes more widely. The forehead shows horizontal lines, and a notable bunching of the skin in the middle could create a small peak or prominence. A slanted line or muscle contraction is visible moving from the inner brow corner to the middle of the forehead, going upward and outward.
	\item Elevating both the inner and outer parts of the eyebrows results in a subtle increase in tension in the skin of the glabella and forehead, leading to a worsening of existing forehead wrinkles and the appearance of new, delicate wrinkles through the forehead’s middle. The eyebrows  exhibit a slightly more arched form, as the outer segments rise and the inner segments are pulled skyward, revealing more of the eyes. The forehead is marked by horizontal creases, and there could be an obvious clustering of skin at the center, sometimes creating a small bump or ridge. A diagonal fold or muscle contraction is seen starting from the inner brow corner and stretching toward the forehead’s midpoint, in an upward and outward direction.
	\item There is a simultaneous lifting of the inner and outer corners of the eyebrows, causing the skin in the brow area and on the forehead to become somewhat tauter. This tautness contributes to the enhancement of forehead wrinkles that are already present and the development of new, subtle wrinkles across the forehead’s center. The eyebrows  now have a slightly more arched silhouette, with the outer sections higher and the inner sections pulled up, making the eyes appear more open. Horizontal marks are visible on the forehead, and there is often a clear gathering of skin in the central part, potentially forming a minor peak or elevation. A diagonal crease or muscle contraction is noticed starting from the inner brow corner and extending up and out to the middle of the forehead.
    \end{enumerate}
        
    \item AU10, AU15 and AU17:
        \begin{enumerate}
        \item The upper lip is raised with the center portion elevated more than the outer portions, creating an angular bend in the upper lip’s shape. The infraorbital triangle is pushed up, deepening the nasolabial furrow. The lip corners are pulled down with a lateral pulling and angling down, horizontally stretching the lips and altering their shape at the corners. The chin boss is severely wrinkled due to being pushed up, and the lower lip is pushed up and out, potentially causing a depression under the lower lip and an inverted-U shape of the mouth. The skin below the lips  show pouching, bagging, or wrinkling
	\item The central part of the upper lip is lifted higher than its sides, causing an angular contour in the upper lip’s form. The infraorbital triangle is elevated, which intensifies the nasolabial fold. The corners of the lips are drawn downward with a sideward tension and a downward angle, horizontally lengthening the lips and modifying their appearance at the corners. The chin prominence becomes heavily creased as it is forced upward, and the lower lip is elevated and protruded, possibly leading to a indentation beneath it and an inverted-U configuration of the mouth. The area beneath the lips might exhibit sagging, bulging, or creasing.
	\item The upper lip is pulled upwards, with the middle section more elevated than the sides, forming an angular deformation in the upper lip. The infraorbital triangle moves upward, deepening the nasolabial crease. The corners of the lips are dragged downward and outward, stretching the lips horizontally and changing their shape at the corners. The chin’s projecting area is heavily wrinkled due to upward pressure, and the lower lip is thrust upward and outward, potentially creating a sunken area under it and an inverted-U shaped mouth. The skin under the lips can show sagging, puffiness, or creases.
	\item Elevating the upper lip with the midsection higher than the lateral parts results in an angular bend to the upper lip’s outline. The infraorbital triangle is thrust upward, which accentuates the nasolabial fold. The lip corners are drawn down with a horizontal pull and a downward slant, elongating the lips laterally and altering their form at the corners. The chin’s protruding area is significantly wrinkled from the upward movement, and the lower lip is pushed up and forward, possibly leading to a concave area beneath it and an inverted-U shaped oral cavity. The region below the lips  appear puffy, baggy, or lined.
	\item The upper lip is lifted in such a way that the middle is more raised than the edges, causing an angular modification to the upper lip’s shape. The infraorbital triangle is compressed upward, enhancing the nasolabial crease. The lips’ corners are pulled downward with a sideward tension and a descending angle, causing the lips to stretch horizontally and altering their appearance at the corners. The protruding part of the chin becomes heavily wrinkled due to the upward force, and the lower lip is lifted and extended, which might result in a depression under it and an inverted-U shaped mouth. The skin beneath the lips could exhibit sagging, bulging, or wrinkling.
    \end{enumerate}
        
    \item AU6, AU7 and AU12:
        \begin{enumerate}
        \item The cheeks are lifted, raising the infraorbital triangle and possibly deepening the infraorbital furrow. The lower eyelid is notably raised and straightened, with a bulge indicating tension and a marked narrowing of the eye aperture. The corners of the lips are pulled back and upward, creating an oblique shape to the mouth and deepening the nasolabial furrow. The skin below the lower eyelid is pushed up, potentially causing wrinkling or bagging, and the upper lip is raised slightly. The overall effect is a complex movement of the facial muscles that results in a significant lifting of the cheeks, lips, and lower lid, with accompanying changes in the skin texture around the eyes and mouth.
	\item The infraorbital triangle is elevated as the cheeks are lifted, which might lead to an increase in the depth of the infraorbital furrow. The lower eyelid is significantly lifted and flattened, showing a bulge that signifies tension and a clear reduction in the size of the eye opening. The mouth’s corners are retracted and pulled skyward, forming an oblique mouth shape and enhancing the nasolabial crease. The skin under the lower eyelid is pushed upward, possibly leading to wrinkling or puffiness, and the upper lip could experience a slight elevation. The collective motion of the facial muscles produces a pronounced lifting of the cheeks, lips, and lower eyelid, along with visible changes in the skin’s texture around the orbital and oral areas.
	\item As the cheeks are hoisted, the infraorbital triangle ascends, which could intensify the infraorbital furrow. There is a noticeable lifting and straightening of the lower eyelid, with a bulge suggesting tension and a distinct narrowing of the eye slit. The mouth corners are drawn backward and upward, giving the mouth an oblique appearance and deepening the nasolabial crease. The area beneath the lower eyelid is shoved up, potentially leading to creases or bags, and the upper lip might also be slightly lifted. The net result is a intricate facial muscle action that yields a substantial elevation of the cheeks, lips, and lower eyelid, accompanied by alterations in the skin’s texture surrounding the eyes and mouth.
	\item With the cheeks raised, the infraorbital triangle is lifted, potentially causing the infraorbital furrow to become more pronounced. The lower eyelid is notably lifted and made taut, with a bulge indicating strain and a significant constriction of the eye’s aperture. The corners of the lips are pulled backward and upward, resulting in an angled mouth shape and a deepened nasolabial crease. The skin below the lower eyelid is forced upward, which  cause wrinkling or sagging, and the upper lip could see a slight rise. The overall movement of the facial muscles creates a notable lifting effect on the cheeks, lips, and lower eyelid, with concurrent changes in the skin texture near the eyes and mouth.
	\item Lifting the cheeks causes the infraorbital triangle to rise, which could contribute to a more pronounced infraorbital furrow. The lower eyelid is significantly lifted and tensed, with a bulge that points to tension and a notable decrease in the width of the eye opening. The lips’ corners are retracted and elevated, forming an oblique mouth contour and increasing the depth of the nasolabial crease. The skin under the lower eyelid is pushed up, possibly leading to wrinkling or puffiness, and the upper lip might be slightly elevated as well. The intricate play of facial muscles results in a marked lifting of the cheeks, lips, and lower eyelid, along with associated changes in the skin’s texture around the eyes and mouth.
    \end{enumerate}
        
    \item AU6, AU12 and AU15:
        \begin{enumerate}
        \item The cheeks are raised, causing the skin below the eyes to become more convex and the infraorbital furrows to deepen. The corners of the mouth are lifted and pulled outwards, but not too much. The upper lip is slightly raised, and the lower lip  protrude slightly. The skin around the mouth is tense, with lines extending from the nose to the corners of the mouth becoming more pronounced.
	\item Elevating the cheeks leads to a more pronounced convexity of the skin beneath the eyes and an increase in the depth of the infraorbital furrows. The mouth’s corners are slightly elevated and drawn to the sides, but not excessively so. There might be a slight upward movement of the upper lip, while the lower lip could exhibit a minor outward projection. The area around the mouth is tight, causing the lines running from the nose to the mouth’s corners to become more visible.
	\item As the cheeks are lifted, the area under the eyes takes on a more convex shape, and the infraorbital furrows become more accentuated. The mouth’s corners are slightly raised and pulled outward, though not to a large degree. The upper lip could show a small lift, and the lower lip might stick out a bit. The skin surrounding the mouth is under tension, making the lines from the nose to the mouth’s corners more distinct.
	\item With the cheeks hoisted up, the skin below the eyes assumes a convex form, and the infraorbital furrows grow deeper. The mouth’s corners are slightly elevated and extended laterally, yet not overly so. The upper lip might be marginally lifted, and the lower lip could have a slight protrusion. The skin in the mouth area is taut, leading to the lines from the nose to the mouth’s corners becoming more marked.
	\item A lifting of the cheeks results in the skin under the eyes becoming more convex and the infraorbital furrows becoming more pronounced. The corners of the mouth experience a slight upward and outward pull, though not to an extreme. The upper lip could see a subtle rise, and the lower lip might be slightly more prominent. The tension in the skin around the mouth causes the lines from the nose to the mouth’s corners to stand out more.
    \end{enumerate}
        
    \item AU12, AU15 and AU17:
        \begin{enumerate}
	\item Lifting the corners of the mouth is accompanied by a noticeable raising of the chin, which causes the lower lip to be pushed upwards and outwards more significantly. The skin below the lower lip is stretched, creating a depression directly under the lip. The chin area is tense, with pronounced wrinkling evident.
	\item A distinct elevation of the mouth’s corners and a marked lifting of the chin lead to a more pronounced pushing up and out of the lower lip. The skin beneath the lower lip stretches, forming an indentation below it. The region of the chin is under tension, displaying noticeable wrinkling.
	\item Raising the mouth’s corners is paired with a clear uplift of the chin, causing the lower lip to move significantly upwards and outwards. The area under the lower lip becomes stretched, resulting in a depression directly beneath the lip. The chin shows tension, characterized by prominent wrinkling.
	\item As the corners of the mouth are lifted, the chin also rises noticeably, which elevates and protrudes the lower lip more than usual. The skin below the lower lip is pulled tight, creating a sunken area under the lip. The chin exhibits tension, with well-defined wrinkling.
	\item There is simultaneous lifting at the corners of the mouth and a notable raising of the chin, which leads to a more accentuated upward and outward movement of the lower lip. The skin below the lower lip stretches, forming a depression directly beneath it. The chin area is tense, with clearly visible wrinkling.

    \end{enumerate}
        
    \item AU12, AU17 and AU23:
        \begin{enumerate}
	\item The corners of the lips are raised and pulled back, resulting in a stretching of the skin around the mouth with the appearance of wrinkles above and below the lips. The lower lip is raised and pushed forward, creating a more pronounced bulge of the lower lip muscle. The chin area is affected by deepening wrinkles as the chin is pushed upward. The overall effect is a narrowing of the lips and a tightening of the facial features around the mouth.
	\item Elevating and retracting the corners of the lips leads to tension in the skin encircling the mouth, manifesting as creases both above and beneath the lips. The lower lip could be lifted and protruded, causing the lower lip muscle to bulge more distinctly. Wrinkles in the chin area intensify as the chin is forced upwards. The net result is a narrowing of the lips and a tensing of the facial muscles surrounding the mouth.
	\item As the lips’ corners are lifted and drawn backwards, the skin near the mouth stretches, forming wrinkles both over and under the lips. The lower lip might also be elevated and extended, leading to a more noticeable protrusion of the lower lip muscle. The chin region is marked by deeper wrinkles due to the upward push of the chin. The overall outcome is a narrowing of the lips and a firming of the facial features around the mouth.
	\item With the lips’ corners pulled back and upwards, the skin around the mouth experiences tension, resulting in the formation of wrinkles that extend above and below the lips. The lower lip could be lifted to project forward, accentuating the bulge of the lower lip muscle. The chin becomes wrinkled more deeply as it is moved upwards. This creates an overall effect of the lips narrowing and the facial features around the mouth becoming tighter.
	\item The act of lifting and pulling back the corners of the lips stretches the surrounding mouth skin, causing wrinkles to appear above and below the lips. The lower lip  rise and jut out, enhancing the prominence of the lower lip muscle. The chin area is affected by an increase in wrinkling due to the upward movement of the chin. The collective impact is a thinner appearance of the lips and a tenser look of the facial muscles in the mouth area.
    \end{enumerate}
        
    \item AU10, AU17 and AU23:
        \begin{enumerate}
	\item The upper lip is drawn straight up, with the center elevated more than the outer portions. The infraorbital triangle is pushed up, and the nasolabial furrow is deepened. The chin boss shows severe wrinkling as it is pushed up, and the lower lip is pushed up and out. The lips are tightened, appearing more narrow, with small wrinkles in the skin above and below the lips, and muscle bulges below the lower lip. The red parts of the lips  roll inwards, almost disappearing from view, while the lips  protrude outward.
	\item The upper lip is pulled directly upwards, with its central part rising higher than the sides. The infraorbital triangle is elevated, causing the nasolabial furrow to become more pronounced. The chin prominence displays significant wrinkling due to its upward movement, and the lower lip is lifted and extended. The lips are tensed, appearing narrower, and are accompanied by fine wrinkles on the skin both above and below the lips, as well as muscle bulges beneath the lower lip. The red portion of the lips  turn inward, nearly vanishing from sight, while the lips could also jut outwards.
	\item The upper lip moves straight upwards, with the middle section lifted more than the edges. The infraorbital triangle is forced upwards, deepening the nasolabial furrow. The chin’s boss shows intense wrinkling as it moves up, and the lower lip is lifted and pushed forward. The lips become tight and seem narrower, with tiny wrinkles on the skin above and below the lips and bulges of muscle below the lower lip. The red areas of the lips might roll in, almost concealed, and the lips could stick out.
	\item Pulling the upper lip upwards results in the center being more elevated than the outer parts. The infraorbital triangle is thrust upwards, which deepens the nasolabial furrow. The chin’s boss is marked by severe wrinkling due to its upward displacement, and the lower lip is elevated and projected outwards. The lips are compressed, taking on a narrower look, with minor wrinkles on the skin above and below the lips and noticeable muscle bulges below the lower lip. The red parts of the lips  curl inwards, nearly disappearing, while the lips themselves  protrude.
	\item The upper lip is raised vertically, with its central area higher than its lateral parts. The infraorbital triangle is lifted, leading to a deepening of the nasolabial furrow. The chin’s boss exhibits pronounced wrinkling as it is elevated, and the lower lip is pushed upwards and outwards. The lips are tightened, appearing to be more slender, with fine lines on the skin above and below the lips and a bulge of muscle beneath the lower lip. The red sections of the lips might roll inwards, almost entirely hidden, and the lips could extend outwards.

    \end{enumerate}
        
    \item AU1 and AU4:
        \begin{enumerate}
	\item A subtle lifting of the inner corners of the eyebrows, with the skin between the eyebrows and on the forehead slightly elevated.  This results in a deepening of wrinkles in the center of the forehead, particularly vertical lines between the eyebrows. The inner parts of the eyebrows  show a slight downward pull at the sides, with possible wrinkling at the corners.  The overall shape of the eyebrows forms an oblique angle, with the medial portion raised and the lateral portion relatively lower.  The upper eyelid and eye cover fold are pulled medially and upwards, creating a triangular shape if visible. The forehead exhibits horizontal wrinkling, primarily in the center, which is curved or form an omega shape.  The area between the eyebrows  show vertical lines, wrinkles, or bunching of skin, with the inner brow corners lifted as the centers are pulled downwards, potentially creating an oblique wrinkle or bulge in the mid-forehead to the inner brow corner region.
	\item A slight elevation of the inner corners of the eyebrows is observed, accompanied by a gentle raising of the skin in the glabella and on the forehead. This action intensifies the wrinkles in the middle of the forehead, especially the vertical lines between the brows. The inner sections of the eyebrows might exhibit a minor downward tension at the edges, possibly accompanied by creases at the corners. The eyebrows’ overall form creates an angled appearance, with the central section higher and the outer sections relatively lower. The upper eyelid and the fold of the eye cover are drawn towards the center and upwards, forming a triangular shape when visible. The forehead displays horizontal creases, mainly centralized, which might be curved or resemble an omega shape. The area between the eyebrows  reveal vertical lines, creases, or skin gathering, as the inner brow corners are lifted while the centers are drawn downwards, possibly leading to an oblique wrinkle or bulge from the mid-forehead to the inner brow corner area.
	\item A delicate upward movement of the inner eyebrow corners is noted, with a concurrent slight rise in the skin of the glabella and forehead. This movement accentuates the central forehead wrinkles, particularly the vertical ones between the eyebrows. There might be a slight pulling down at the sides of the inner eyebrows, potentially with some wrinkling at the corners. The eyebrows’ shape takes on an oblique angle, with the medial sections higher and the lateral sections lower. The upper eyelid and the eye cover fold are tugged medially and upwards, possibly revealing a triangular shape. The forehead shows horizontal creases, focused centrally, which can be curved or omega-shaped. The space between the eyebrows  present vertical lines, creases, or skin folding, as the inner brow corners are lifted while the centers pull downwards, possibly causing an oblique wrinkle or bulge from the middle of the forehead to the inner brow corner.
	\item A small lift of the inner parts of the eyebrows is seen, with a corresponding minor lifting of the skin in the area between the eyebrows and on the forehead. This action deepens the central forehead wrinkles, especially the vertical lines in the glabella. The inner sections of the eyebrows might show a slight downward draw at their sides, with potential wrinkling at the corners. The eyebrows assume an oblique shape, with the middle sections higher and the outer sections lower. The upper eyelid and the fold above the eye are pulled towards the center and upwards, which could expose a triangular form. The forehead is marked by horizontal creases, most noticeable in the center, which might be arched or shaped like an omega. The region between the eyebrows  exhibit vertical lines, creases, or skin bunching, as the inner brow corners are elevated while the centers pull down, possibly forming an oblique wrinkle or bulge from the mid-forehead to the inner brow corner.
	\item A subtle raising of the inner corners of the eyebrows occurs, along with a slight uplifting of the skin in the glabella and on the forehead. This results in more pronounced wrinkles in the middle of the forehead, particularly the vertical lines between the eyebrows. The inner parts of the eyebrows  experience a slight downward tension at the sides, possibly with wrinkling at the corners. The eyebrows’ shape takes on an oblique angle, with the medial sections higher and the lateral sections relatively lower. The upper eyelid and the fold of the eye cover are drawn towards the center and upwards, potentially revealing a triangular shape. The forehead is characterized by horizontal creases, primarily in the center, which might be curved or form an omega shape. The space between the eyebrows  show vertical lines, creases, or skin gathering, as the inner brow corners are lifted while the centers are pulled downwards, possibly creating an oblique wrinkle or bulge from the mid-forehead to the inner brow corner area.
    \end{enumerate}
        
    \item AU1 and AU2:
        \begin{enumerate}
	\item A noticeable lifting of the inner and outer portions of the eyebrows, with the inner corners raised slightly and the outer parts elevated more pronouncedly. The skin between the eyebrows and on the forehead is lifted, causing slight deepening of existing wrinkles and the formation of new ones, particularly in the center of the forehead. An upward movement of the eyebrows, creating an arched and curved appearance. The forehead skin is gathered, leading to the emergence of horizontal wrinkles across the entire forehead. The eye cover fold is stretched, becoming more visible, and in some cases, it reveals the upper eyelid, especially in individuals with deeply set eyes.
	\item Both the inner and outer sections of the eyebrows are visibly lifted, with the inner corners undergoing a slight elevation and the outer sections a more distinct lift. The area between the eyebrows and the forehead skin is raised, resulting in a slight increase in the depth of current wrinkles and the creation of new ones, especially at the forehead’s center. The eyebrows move upwards, forming an arched and curved shape. The forehead skin is drawn together, causing horizontal wrinkles to appear across the entire forehead. The fold of the eye cover is extended, making it more noticeable, and in certain cases, it exposes the upper eyelid, particularly in people with deeply sunken eyes.
	\item A clear lifting of the inner and outer segments of the eyebrows is observed, where the inner corners are slightly uplifted and the outer segments are more significantly raised. The lifting of the skin in the glabella and on the forehead leads to a mild intensification of existing wrinkles and the appearance of new ones, particularly in the middle of the forehead. The eyebrows are raised, giving them an arched and curved look. The forehead skin is pulled together, resulting in the development of horizontal wrinkles that span the width of the forehead. The eye cover fold becomes more apparent due to stretching, sometimes revealing the upper eyelid, especially in those with deep-set eyes.
	\item There is a distinct lifting of both the inner and outer parts of the eyebrows, with a mild raise at the inner corners and a more obvious lift at the outer parts. The skin in the glabella and on the forehead is lifted, causing a slight deepening of old wrinkles and the formation of new ones, mainly in the central forehead area. The eyebrows are pulled upwards, creating an arched and curved effect. The forehead skin is gathered, which brings about horizontal wrinkles across the entire forehead. The eye cover fold stretches and becomes more visible, in some instances showing the upper eyelid, especially in individuals with eyes that are set deeply.
	\item A notable lifting occurs at the inner and outer regions of the eyebrows, with the inner corners experiencing a minor lift and the outer regions a more accentuated one. The skin between the eyebrows and on the forehead is raised, leading to a slight enhancement of current wrinkles and the onset of new ones, particularly centralized on the forehead. The eyebrows are elevated, producing an arched and curved appearance. The forehead skin is drawn together, causing horizontal wrinkles to emerge across the full width of the forehead. The eye cover fold is elongated and more noticeable, sometimes uncovering the upper eyelid, especially in people whose eyes are deeply embedded.

    \end{enumerate}
        
    \item AU10 and AU15:
        \begin{enumerate}
	\item The center of the upper lip is drawn straight up with a more subtle elevation of the outer portions, creating an angular bend in the upper lip’s shape. The infraorbital triangle is pushed up, deepening the nasolabial furrow, which  already is evident in a neutral expression. The lip corners are simultaneously pulled down slightly, with a lateral pulling and angling down of the corners, resulting in a stretching of the lips horizontally. Below the lip corners, there is some pouching, bagging, or wrinkling of the skin, which is more pronounced with a stronger action. The chin boss  show flattening or the appearance of bulges, and a depression could form medially under the lower lip.
	\item The central part of the upper lip is pulled upwards vertically while the outer sections are elevated less dramatically, forming an angular contour in the upper lip’s outline. The infraorbital triangle is lifted, exacerbating the nasolabial fold, which might already be noticeable in a resting expression. Concurrently, the lip corners are slightly lowered, with a sideward tug and a downward angle at the corners, leading to a horizontal stretching of the lips. Beneath the lip corners, there is a slight pouching, sagging, or creasing of the skin, which can become more accentuated with a more intense action. The chin prominence  appear flattened or develop bulges, and a indentation might occur in the center below the lower lip.
	\item An upward pull on the middle of the upper lip, with a less obvious lift on the sides, forms an angular shape in the upper lip. The area of the infraorbital triangle is raised, increasing the depth of the nasolabial folds, which might be visible even at rest. The corners of the lips are gently drawn downwards, with a sideways and downward tension that stretches the lips laterally. Below the lip corners, the skin might pucker, sag, or crease, a feature that can become more distinct with a forceful expression. The chin’s fullness  seem to flatten or present with bulges, and a central dip might form beneath the lower lip.
	\item The middle of the upper lip is pulled tautly upwards with a gentler lifting of the lip’s outer edges, creating an angular bend in the upper lip’s form. The infraorbital triangle is pushed upwards, deepening the nasolabial crease, which is often noticeable in a neutral facial expression. The lip corners are slightly pulled downwards and outwards, causing a horizontal extension of the lips. Below the corners of the lips, there is some bulging, sagging, or wrinkling of the skin, which can be more evident with a stronger facial movement. The chin’s prominence might appear to flatten or develop bulges, and a central depression could form under the lower lip.
	\item A direct upward movement of the upper lip’s center, contrasted with a milder lift at the sides, results in an angular modification to the upper lip’s contour. The infraorbital triangle experiences an upward force, which intensifies the nasolabial folds, potentially visible in a relaxed state. The corners of the lips are concurrently drawn slightly down and outward, leading to a lateral stretching of the lips. The skin under the lip corners  exhibit some puffiness, droopiness, or creasing, which can be more pronounced with a forceful action. The chin’s boss  seem to flatten or have bulges, and a central indentation might appear below the lower lip.
    \end{enumerate}
        
    \item AU10 and AU17:
        \begin{enumerate}
	\item The upper lip is raised with the center portion elevated more than the outer portions, causing an angular bend in the upper lip’s shape. The infraorbital triangle is pushed up, deepening the nasolabial furrow. The chin boss exhibits severe wrinkling as it is pushed up, and the lower lip is pushed up and out markedly. The shape of the mouth  appear more pronounced with an inverted-U shape, and the lower lip  protrude if the action is strong.
	\item The central part of the upper lip is lifted higher than its sides, creating an angular contour in the upper lip’s form. The infraorbital triangle is elevated, which intensifies the nasolabial fold. The chin boss becomes heavily wrinkled due to being raised, and the lower lip is significantly lifted and pushed outward. The mouth  take on a more distinct inverted-U shape, and the lower lip could stick out if the movement is forceful.
	\item Raising the upper lip with a greater uplift at the center than at the edges results in an angular deformation of the upper lip. The infraorbital triangle is forced upward, enhancing the nasolabial furrow. The chin boss is subjected to pronounced wrinkling as it moves up, and the lower lip is distinctly raised and thrust forward. The mouth might adopt a more accentuated inverted-U form, and the lower lip could become more prominent with a strong action.
	\item The center of the upper lip is pulled up more than the lateral parts, causing an angular bend to the upper lip’s outline. The infraorbital triangle is lifted, thereby deepening the nasolabial crease. The chin boss shows deep wrinkles as it is elevated, and the lower lip is visibly lifted and extended. The mouth’s shape can become more noticeable with an inverted-U configuration, and the lower lip  jut out if the expression is intense.
	\item Elevating the upper lip with an emphasis on the central area relative to the sides leads to an angular change in the upper lip’s shape. The infraorbital triangle is raised, which contributes to a deeper nasolabial furrow. The chin boss experiences significant wrinkling as it is pushed upwards, and the lower lip is distinctly pushed up and out. The configuration of the mouth is more evident with an inverted-U form, and the lower lip could extend outward with a vigorous action.
    \end{enumerate}    
    
    \item AU15 and AU17:
        \begin{enumerate}
	\item The lip corners are pulled down slightly with a lateral pulling and angling down, stretching the lips horizontally and changing their shape at the corners. Below the lips, the skin  show pouching, bagging, or wrinkling, and the chin boss is flattened or show bulges. The chin boss is pushed up severely, causing extreme wrinkling, and the lower lip is pushed up and out, which result in a depression medially under the lower lip and an inverted-U shape of the mouth.
	\item The corners of the lips are gently drawn downwards with a sideward tug and a downward angle, which lengthens the lips horizontally and alters their form at the corners. The skin beneath the lips might exhibit some sagging, puffiness, or creases, and the chin boss could appear flattened or have protrusions. The chin boss is significantly elevated, leading to intense wrinkling, while the lower lip is lifted and protruded, potentially creating a central indentation under the lower lip and an inverted-U mouth shape.
	\item Slight downward traction on the lip corners, accompanied by a lateral pull and a downturn, extends the lips horizontally and modifies their appearance at the corners. The area under the lips might show signs of sagging, swelling, or creases, and the chin boss might become flattened or develop bulges. A marked upward push on the chin boss results in severe wrinkling, and the lower lip is thrust upwards and outwards, possibly leading to a depression in the middle beneath the lower lip and an inverted-U shaped mouth.
	\item A minor downward pull on the lip corners, with a sideward component and a downward slant, stretches the lips laterally and changes their shape at the edges. Beneath the lips, the skin could appear puffy, baggy, or wrinkled, and the chin boss might look compressed or have bulges. A strong upward force on the chin boss causes profound wrinkling, and the lower lip is pushed up and out, which can create a central dip under the lower lip and an inverted-U configuration of the mouth.
	\item The lips’ corners are slightly lowered with a dragging motion to the sides and a downward tilt, causing the lips to stretch horizontally and altering their shape at the corners. The skin under the lips might display pouching, sagging, or creases, and the chin boss could either be flattened or exhibit bulges. A forceful upward movement of the chin boss leads to extensive wrinkling, while the lower lip is elevated and pushed forward, possibly resulting in a medial depression below the lower lip and an inverted-U shaped mouth.
    \end{enumerate}
        
    \item AU10 and AU14:
        \begin{enumerate}
	\item The upper lip is elevated with the center portion raised more significantly than the sides, resulting in a straightening of the upper lip line. The infraorbital area is pushed upward, leading to a deepening of the nasolabial folds. Concurrently, the corners of the mouth are drawn tightly inward, causing the lips to narrow at the corners. The skin around the lips is pulled taut, creating pronounced wrinkles and possibly a bulge at the lip corners. Additionally, a deep dimple-like wrinkle  extend beyond the corners of the mouth. The overall shape of the lips tends to be straight rather than curved, and the skin below the lip corners and the chin area is pulled up, contributing to a flattening and stretching of the lower facial skin.
	\item The center of the upper lip rises more prominently than its edges, causing the upper lip line to become straight. As the infraorbital region is lifted, the nasolabial folds become more pronounced. Simultaneously, the mouth corners are pulled tightly towards the center, narrowing the lips at their edges. The skin around the lips is stretched, resulting in distinct wrinkles and potentially a bulge at the lip corners. Moreover, a deep dimple-like wrinkle might extend past the mouth corners. The lips’ general shape becomes more linear than curved, and the skin beneath the lip corners and the chin is lifted, leading to a flattening and extension of the lower facial skin.
	\item With the central part of the upper lip lifted more than the sides, the upper lip line becomes straighter. The infraorbital area’s upward push deepens the nasolabial folds. Meanwhile, the mouth corners are drawn tightly inwards, making the lips appear narrower at the corners. The skin around the lips is pulled taught, which accentuates wrinkles and  form a bulge at the lip corners. Furthermore, a significant dimple-like wrinkle could extend from the mouth corners. The lips assume a straighter form rather than a curved one, and the skin under the lip corners and the chin is elevated, causing the lower facial skin to flatten and stretch.
	\item An elevation of the upper lip with a more pronounced rise in the center than at the sides straightens the upper lip’s contour. The infraorbital region’s upward movement accentuates the nasolabial folds. At the same time, the mouth corners are pulled in tightly, causing the lips to slim at the corners. The skin surrounding the lips is stretched tightly, leading to well-defined wrinkles and possibly a bulge forming at the lip corners. Additionally, a deep wrinkle resembling a dimple  appear beyond the mouth corners. The lips’ shape tends towards straight rather than curved, and the skin below the lip corners and on the chin is lifted, contributing to the flattening and stretching of the lower facial skin.
	\item The upper lip is raised, with the middle section showing a greater increase in height than the side sections, leading to a straightening effect on the upper lip’s line. The infraorbital area experiences an upward force, which in turn deepens the nasolabial folds. In conjunction, the mouth corners are pulled snugly towards each other, narrowing the lips at the corners. The skin around the lips is pulled taut, which highlights wrinkles and could create a bulge at the lip corners. Also, a pronounced dimple-like wrinkle might extend past the mouth corners. The overall shape of the lips tends to be more straight than curved, and the skin under the lip corners as well as the chin area is lifted, resulting in a flattening and stretching of the skin on the lower face.
    \end{enumerate}


    \item AU14 and AU17:
    \begin{enumerate}
	\item The mouth corners are pulled back and upwards with significant tension, resulting in a marked narrowing and straightening of the lips. The skin around the lip corners is severely wrinkled due to the intense inward pull, and the lower lip  be stretched and flattened against the upper lip. Additionally, the chin and lower lip area are subjected to an upward force, which causes the skin on the chin to wrinkle severely and the lower lip to be pushed upwards, further narrowing the red part of the lips. This upward movement  also create a depression below the center of the lower lip. The overall effect is a tight, compressed appearance of the lips with pronounced wrinkling at the corners and on the chin, and a distinctive change in the typical shape of the lips.
	\item The corners of the mouth are retracted and elevated with considerable strain, leading to a pronounced thinning and straightening of the lips. The skin at the lip corners is heavily creased because of the strong inward traction, and the lower lip might be elongated and pressed against the upper lip. Furthermore, an upward pressure is applied to the chin and lower lip region, causing the chin’s skin to wrinkle deeply and the lower lip to be elevated,narrowing the vermillion of the lips even more. This upward lifting can also result in a concave area beneath the midpoint of the lower lip. The net result is a taut, compact look of the lips with obvious creasing at the corners and on the chin, along with a marked alteration to the usual lip shape.
	\item Drawing the mouth corners back and up with considerable force leads to a significant tightening and straightening of the lip line. The area around the lip corners becomes deeply wrinkled due to the intense inward pull, while the lower lip  get stretched and pressed against the upper lip. Also, the chin and the region of the lower lip experience an upward push, causing the chin’s skin to wrinkle intensely and the lower lip to be raised, thus narrowing the red part of the lips further. This action might also induce a depression to form under the lower lip’s center. The overall outcome is a lips appearance that is tight and compressed, with pronounced wrinkling at the corners and on the chin, and a clear transformation of the lips’ normal contour.
	\item A significant backward and upward pull on the mouth corners creates a noticeable narrowing and linear appearance of the lips. The skin surrounding the lip corners is severely wrinkled as a result of the forceful inward drawing, and the lower lip could be drawn out and flattened against the upper lip. In addition, an upward force is exerted on the chin and the lower lip area, leading to severe wrinkling of the chin’s skin and an upward push of the lower lip, which narrows the red portion of the lips even more. This upward movement might also cause a indentation to appear below the central part of the lower lip. The collective effect is a lips look that is tense and compressed, with well-defined wrinkling at the corners and on the chin, and a distinct shift from the typical shape of the lips.
	\item There is a pronounced retraction and elevation of the mouth corners, causing the lips to become notably slender and straight. The skin near the lip corners is heavily creased due to the strong inward traction, and the lower lip might be elongated and pressed flat against the upper lip. Moreover, an upward pressure is applied to the chin and lower lip regions, resulting in the chin’s skin becoming deeply wrinkled and the lower lip being elevated, thus further narrowing the vermillion of the lips. This upward action could also give rise to a sunken area below the middle of the lower lip. The overall impact is a lips appearance that is tight and compressed, with prominent creasing at the corners and on the chin, and a clear modification to the standard shape of the lips.
    \end{enumerate}
        
    \item AU6 and AU12:
        \begin{enumerate}
	\item The cheeks are lifted, raising the infraorbital triangle and slightly deepening the infraorbital furrow. The corners of the lips are pulled back and upward, creating an oblique angle to the mouth and deepening the nasolabial furrow. The skin adjacent to the nasolabial furrow is raised up and laterally. The lower eyelid  show signs of bagging or wrinkling, and the eye aperture  be slightly narrowed due to the upward push of the cheek and skin below the lower lid.
	\item Elevating the cheeks results in the infraorbital triangle being raised and a minor increase in the depth of the infraorbital furrow. The mouth corners are retracted and elevated, forming an oblique mouth angle and intensifying the nasolabial crease. The skin near the nasolabial fold is lifted and pulled to the side. The lower eyelid might exhibit sagging or creases, and the eye opening could narrow slightly due to the upward displacement of the cheek and the skin beneath the lower eyelid.
	\item As the cheeks are hoisted, the infraorbital triangle ascends, slightly intensifying the infraorbital crease in the process. The lips’ corners are drawn backwards and upwards, leading to an angled mouth appearance and a more pronounced nasolabial fold. The tissue beside the nasolabial fold is lifted and directed laterally. The lower eyelid could display puffiness or wrinkles, and the eye’s width  decrease because of the upward force exerted by the cheek and the skin under the lower eyelid.
	\item The lifting of the cheeks causes an elevation of the infraorbital triangle while also marginally increasing the depth of the infraorbital crease. The mouth corners are pulled backwards and upwards, causing the mouth to form an oblique shape and deepening the nasolabial fold. The skin around the nasolabial fold is raised and shifted to the side. The lower eyelid might show signs of swelling or creasing, and the eye’s opening  become slightly more narrow due to the upward movement of the cheek and the skin below the lower eyelid.
	\item With the cheeks raised, the infraorbital triangle is uplifted, slightly deepening the infraorbital furrow in the process. The corners of the lips are retracted and elevated, producing an oblique orientation of the mouth and enhancing the nasolabial crease. The skin next to the nasolabial fold is lifted and moved laterally. The lower eyelid could exhibit bagging or wrinkling, and the eye’s aperture might be somewhat constricted as a result of the upward thrust from the cheek and the skin under the lower eyelid.

    \end{enumerate}
        
    \item AU7 and AU12:
        \begin{enumerate}
	\item The lower eyelid is raised and straightened, causing a bulge and narrowing the eye aperture. The corners of the lips are markedly raised and angled up obliquely, with a deepened nasolabial furrow. The infraorbital triangle is lifted slightly, and the skin below the lower eyelid is pushed up, potentially creating wrinkles or bags. The overall effect is a pronounced narrowing of the eye aperture and a noticeable lifting of the lower lid and lip corners.
	\item The lower eyelid is raised and straightened, causing a bulge and narrowing the eye aperture. The corners of the lips are markedly raised and angled up obliquely, with a deepened nasolabial furrow. The infraorbital triangle is lifted slightly, and the skin below the lower eyelid is pushed up, potentially creating wrinkles or bags. The overall effect is a pronounced narrowing of the eye aperture and a noticeable lifting of the lower lid and lip corners.
	\item The lower eyelid is pulled up and made straight, which leads to a protuberance and a reduction in the size of the eye opening. The corners of the lips are distinctly elevated and tilted upwards at an oblique angle, with the nasolabial crease becoming more pronounced. A minor lift of the infraorbital triangle occurs, and the skin beneath the lower eyelid might be elevated, potentially causing creases or swelling. The overall impact is a significant narrowing of the eye aperture and a notable lifting of the lower eyelid and the corners of the lips.
	\item Raising the lower eyelid causes it to become taut and form a bulge, thereby constricting the eye’s aperture. The corners of the lips are visibly lifted and directed upwards at an angle, deepening the nasolabial furrow. A slight elevation of the infraorbital triangle is observed, and the skin below the lower eyelid could be pushed upwards, potentially resulting in wrinkles or puffiness. The general effect is a marked narrowing of the eye opening and a striking uplift of the lower eyelid and lip corners.
	\item As the lower eyelid is lifted and aligned straight, it creates a bulge that narrows the eye opening. The corners of the lips are notably raised and slanted obliquely upwards, with the nasolabial furrow becoming more distinct. The infraorbital triangle experiences a slight lift, and the skin under the lower eyelid might be pushed upwards, leading to the appearance of wrinkles or bags. The end result is a pronounced narrowing of the eye aperture and a clear lifting of the lower eyelid and the corners of the lips.
    \end{enumerate}
        
    \item AU12 and AU17:
        \begin{enumerate}
	\item The corners of the lips are notably elevated and angled upwards, with a slight deepening of the lines running from the nose to the mouth corners. The area below the eyes is slightly raised, and the chin area exhibits severe wrinkling as it is pushed upwards, resulting in a marked elevation of the lower lip. The overall appearance of the lips is altered, with the lower lip protruding and the upper lip possibly lifting to reveal less of the upper teeth. The skin below the lower eyelid  show signs of bagging and wrinkling, and the space between the eyelids  narrow due to the upward movement of the cheek tissue. Additionally, wrinkles  appear at the outer corners of the eyes, and the chin area  show a depression directly below the lower lip as the skin is stretched.
	\item A significant upward lift and angling of the lip corners is observed, accompanied by a minor increase in the depth of the lines extending from the nose to the mouth corners. The region beneath the eyes experiences a slight lifting, while the chin area is marked by intense wrinkling due to being pushed upwards, leading to a pronounced lifting of the lower lip. This results in a changed lip appearance, where the lower lip sticks out and the upper lip might rise, exposing less of the upper teeth. The skin under the lower eyelid could exhibit bagging and wrinkling, and the gap between the eyelids  decrease because of the upward shift of the cheek tissue. Moreover, creases might form at the outer edges of the eyes, and a depression could be seen under the lower lip in the chin area as the skin stretches.
	\item There is a clear upward movement and tilt of the mouth corners, with a subtle enhancement of the nasolabial lines. Below the eyes, the area is slightly lifted, and the chin region shows pronounced wrinkling from being raised, causing the lower lip to elevate significantly. The lips’ overall look  transform, with the lower lip becoming more prominent and the upper lip potentially rising to show less of the upper teeth. The lower eyelid’s skin might display signs of puffiness and wrinkling, and the space between the eyelids could narrow due to the upward displacement of the cheek tissue. Additionally, wrinkles could emerge at the outer corners of the eyes, and the chin might feature a depression directly beneath the lower lip as the skin stretches.
	\item An obvious lifting and upward angling of the lip corners is seen, with a small deepening of the lines from the nose to the mouth corners. The area under the eyes is a bit elevated, and the chin experiences severe wrinkling as it moves upwards, causing the lower lip to rise markedly. This can alter the lips’ general appearance, making the lower lip more projecting and the upper lip possibly lifting to conceal more of the upper teeth. The skin below the lower eyelid might show bagging and wrinkling, and the distance between the eyelids could become narrower due to the upward movement of the cheek tissue. Also, creases might appear at the outer sides of the eyes, and the chin could show a sunken area directly below the lower lip as the skin stretches.
	\item A notable lifting and slanting upwards of the lip corners is evident, along with a slight enhancement of the lines that stretch from the nose to the mouth corners. The region below the eyes is slightly lifted, while the chin area is marked by intense wrinkling due to its upward push, which leads to a clear lifting of the lower lip. This change can result in a modified appearance of the lips, where the lower lip sticks out further and the upper lip might elevate to reveal less of the upper teeth. The skin under the lower eyelid could exhibit signs of puffiness and wrinkling, and the space between the eyelids  decrease with the upward shift of the cheek tissue. Furthermore, wrinkles might form at the outer corners of the eyes, and a depression could form directly under the lower lip in the chin area as the skin stretches.

    \end{enumerate}
        
    \item AU12 and AU15:
        \begin{enumerate}
	\item The corners of the mouth are lifted and pulled outwards, creating an angular shape. The muscles around the mouth are tense, resulting in a slight narrowing of the mouth opening. The skin around the nose and mouth is drawn tight, leading to the appearance of lines extending from the nose to the corners of the mouth. The lower lip  show a subtle bulge, and there is a visible wrinkling immediately beyond the lip corners.
	\item The mouth corners are raised and drawn to the sides, forming a more angular contour. The muscles encircling the mouth are tight, causing a minor constriction of the mouth’s aperture. The area around the nose and mouth is pulled taut, causing lines to emerge from the nose towards the mouth’s corners. The lower lip might exhibit a slight protrusion, and there is evident wrinkling right outside the corners of the lips.
	\item Elevating and outwardly stretching the mouth corners produces an angular mouth shape. The muscles near the mouth are under tension, slightly narrowing the space between the lips. The skin surrounding the nose and mouth is stretched, revealing lines that run from the nose to the mouth’s edges. A subtle swell might be seen on the lower lip, and creases are noticeable just past the mouth corners.
	\item An outward and upward pull on the mouth corners shapes them into an angle. Tension in the muscles around the mouth leads to a slight reduction in the width of the mouth opening. The skin near the nose and mouth is pulled taught, causing creases to form from the nose to the mouth corners. The lower lip could display a faint bulge, and there is clear wrinkling adjacent to the lip corners.
	\item The mouth corners are raised and extended laterally, forming a more pointed angle. The muscles that frame the mouth are stiff, which slightly narrows the mouth’s opening. The skin around the nose and mouth is drawn snugly, resulting in lines that stretch from the nose to the mouth’s corners. A slight bulge might be visible on the lower lip, and there is distinct wrinkling right beside the lip corners.
    \end{enumerate}
        
    \item AU12 and AU23:
        \begin{enumerate}
	\item The outer edges of the eyebrows are raised, causing the skin on the forehead to wrinkle horizontally. The eyelids are slightly tensed, and the eyes appear more open. The cheeks are elevated, which  result in a temporary reduction of the visibility of the lower eyelid bags. The lips are stretched horizontally, with the corners of the mouth pulled slightly upward.
	\item The lateral ends of the eyebrows are lifted, leading to horizontal creases forming across the forehead. There is a mild tensing of the eyelids, making the eyes seem wider. The cheeks are raised, potentially lessening the prominence of lower eyelid puffiness. The lips are drawn out horizontally, with a minor upward pull at the mouth’s corners.
	\item Elevating the outer portions of the eyebrows causes the forehead skin to crease from side to side. The eyelids experience a bit of tension, giving the eyes a more open look. As the cheeks are lifted, the visibility of bags under the lower eyelids  decrease temporarily. The lips are stretched laterally, with a subtle lifting of the mouth corners.
	\item The outer tips of the eyebrows are hoisted, resulting in transverse lines on the forehead. The eyelids are slightly tense, contributing to a more awake appearance of the eyes. The cheeks are lifted, which can diminish the appearance of lower eyelid swelling. The lips are pulled sideways, with a slight upward tug at the corners of the mouth.
	\item Raising the outer parts of the eyebrows leads to horizontal wrinkling across the forehead. The eyelids are faintly tensed, causing the eyes to look more alert. The cheeks are elevated, possibly lessening the noticeable bagginess under the lower eyelids. The lips are stretched in a horizontal direction, with a slight upward pull at the mouth’s corners.
    \end{enumerate}    
    \item AU12 and AU24:
        \begin{enumerate}
	\item The lips are pursed together tightly, with the inner edges of the lips pressing against each other and the outer edges remaining relatively unchanged. This action creates a subtle bulge in the center of the lips and  cause vertical wrinkles to form around the lip area. The overall width of the lips appears reduced, and the upper lip  lift slightly, exposing less of the lower teeth.
	\item The lips are clenched firmly, causing the inner parts to press into one another while the outer parts stay mostly the same. This motion results in a slight protrusion at the center of the lips and can lead to the development of vertical creases near the lips. The total breadth of the lips seems diminished, and the upper lip might elevate a bit, revealing less of the lower teeth.
	\item The lips are clenched firmly, causing the inner parts to press into one another while the outer parts stay mostly the same. This motion results in a slight protrusion at the center of the lips and can lead to the development of vertical creases near the lips. The total breadth of the lips seems diminished, and the upper lip might elevate a bit, revealing less of the lower teeth.
	\item A firm pressing of the lips to each other occurs, where the inner lip margins are compressed while the outer margins stay relatively stable. This causes a minute swelling in the center of the lips and can induce the appearance of vertical creases around the lip perimeter. The overall lip width seems to decrease, and the upper lip might be slightly elevated, thus exposing a smaller portion of the lower teeth.
	\item The lips are tightly pursed, with the inner borders of the lips pushing against one another, while the outer borders do not change significantly. This action generates a subtle protrusion at the lips’ center and might result in vertical folds forming around the lips. The width of the lips is seemingly decreased, and the upper lip could elevate a tad, concealing more of the lower teeth.
    \end{enumerate}
        
    \item AU14 and AU23:
        \begin{enumerate}
	\item The facial expression is marked by a pronounced tightening of the lips, with the corners of the mouth pulled tightly inward, resulting in significant wrinkling around the lip area. The lips are narrowed, with the red portions almost vanishing, and the skin on the chin and lower lip is stretched towards the lip corners. The chin boss appears flattened and stretched, and the nasolabial furrows are deepened. Fine wrinkles and lines are visible above and below the lips, although their prominence is lessened due to the lateral pull on the skin.
	\item A distinct tensing of the facial muscles causes the lips to tighten considerably, with the mouth corners drawn snugly towards the center, leading to pronounced creasing around the lips. The lips become slimmed down, nearly obliterating the visible red parts, while the skin on the chin and lower lip is pulled towards the mouth corners. The chin prominence seems flattened and elongated, and the nasolabial folds are more pronounced. Wrinkles and lines are apparent both above and below the lips, yet their emphasis is reduced by the outward pull on the skin.
	\item The face exhibits a strong tightening of the lips, where the mouth corners are retracted sharply inward, causing notable wrinkling in the lip region. The lips appear narrowed, with the red parts of the lips almost disappearing, and the skin on the lower lip and chin is stretched towards the corners of the mouth. The chin’s fullness is compressed and stretched out, and the nasolabial creases become deeper. Fine lines and wrinkles are seen above and below the lips, but their visibility is diminished due to the skin being pulled laterally.
	\item A marked constriction of the lips is evident in the facial expression, with the mouth corners being drawn tightly inwards, leading to substantial wrinkling around the lips. The lips are narrowed to the point where the red portions are almost concealed, and the skin on the lower lip and chin moves towards the mouth corners. The chin’s projection appears flattened and extended, and the nasolabial grooves are more accentuated. Subtle wrinkles and lines are present above and below the lips, though their prominence is lessened by the skin’s sideward tension.
	\item The facial muscles show a pronounced tightening that affects the lips, pulling the corners of the mouth tightly to the inside and causing significant wrinkling around the lip area. The lips become narrow, with the red sections of the lips nearly invisible, and the skin on the chin and lower lip stretches towards the corners of the mouth. The chin’s roundness is flattened and elongated, and the nasolabial folds are deepened. Fine lines and wrinkles are noticeable above and below the lips, yet they are less pronounced due to the skin being drawn laterally.
    \end{enumerate}


    \item AU17 and AU23:
        \begin{enumerate}
	\item The chin area is raised, creating deep vertical wrinkles below the lower lip. The lips are drawn inward tightly, with the red portion of the lips becoming almost invisible. The skin around the lips is heavily wrinkled, and the lower lip is pushed outward while being raised.
	\item Elevating the chin region results in pronounced vertical creases forming beneath the lower lip. The lips are pulled tightly towards the center, causing the red part of the lips to nearly disappear. The area surrounding the lips is marked by significant wrinkling, and the lower lip is thrust outward as it is lifted.
	\item The chin is lifted, causing deep vertical lines to appear under the lower lip. The lips are drawn tightly inward, making the red portion of the lips almost indiscernible. Wrinkles are abundant around the lips, and the lower lip is both elevated and pushed forward.
	\item As the chin is raised, it produces deep vertical wrinkles below the lower lip. The lips are clenched tightly, nearly hiding the red of the lips. The skin encircling the lips is heavily creased, and the lower lip is simultaneously elevated and protruded.
	\item A lifting of the chin area generates deep vertical folds beneath the lower lip. The lips are tightly retracted, with the red lip area almost concealed. The tissue around the lips is heavily wrinkled, and the lower lip experiences an outward push as it is raised.
    \end{enumerate}
        
    \item AU17 and AU24:
        \begin{enumerate}
	\item The chin is elevated, causing pronounced vertical lines to form on the chin. The lips are compressed tightly together, with the lower lip being pushed up and out. The upper lip  show small wrinkles, and there is a noticeable bulging of the skin above the upper lip and below the lower lip.
	\item An elevation of the chin results in the appearance of distinct vertical creases on the chin area. The lips are squeezed firmly, causing the lower lip to be raised and protruded. The upper lip might exhibit minor wrinkling, and a notable protrusion of the skin is visible both above the upper lip and beneath the lower lip.
	\item Raising the chin leads to the development of clear vertical lines on the chin. The lips are pressed tightly against each other, with the lower lip being lifted and extended outward. Wrinkles  become visible on the upper lip, and there is an obvious bulge in the skin located above the upper lip and below the lower lip.
	\item As the chin is lifted, it causes the formation of well-defined vertical lines across the chin. The lips are compressed with force, pushing the lower lip upwards and outwards. The upper lip could have slight wrinkles, and a discernible swelling of the skin appears over the upper lip and under the lower lip.
	\item The chin’s upward movement creates pronounced vertical folds on the chin. The lips are tightly compressed, which forces the lower lip to move upwards and out. Fine wrinkles might be seen on the upper lip, and a prominent bulge in the skin is evident above the upper lip as well as below the lower lip.

    \end{enumerate}
        
    \item AU15 and AU23:
        \begin{enumerate}
	\item The corners of the mouth move downward slightly, the lips are tightened and drawn inward, reducing the visibility of the red portion of the lips. The skin around the lips becomes more wrinkled.
	\item A slight downturn occurs at the corners of the mouth, while the lips are tensed and pulled towards the center, diminishing the red part of the lips’ visibility. The area surrounding the lips exhibits increased wrinkling.
	\item The mouth corners dip slightly downward, with the lips contracting and retracting, which lessens the prominence of the red sections of the lips. The skin in the lip area shows more creases.
	\item A minimal descent of the mouth corners is accompanied by the lips’ tightening and inward movement, concealing the red part of the lips. The skin around the lips is more wrinkled as a result.
	\item As the corners of the mouth edge slightly down, the lips are tightened and pulled inwards, making the red portion of the lips less visible. Wrinkles around the lips become more pronounced.
    \end{enumerate}
        
\end{itemize}

\end{document}